\pdfoutput=1

\documentclass[11pt]{article}

\usepackage[final]{acl}

\usepackage{times}
\usepackage{latexsym}
\usepackage{multirow}
\usepackage{booktabs}
\usepackage[table]{xcolor} 
\definecolor{Gray}{gray}{0.9} 
\usepackage{graphicx}
\usepackage{subcaption}

\usepackage{adjustbox}
\usepackage[T1]{fontenc}

\usepackage[utf8]{inputenc}

\usepackage{microtype}

\usepackage{inconsolata}

\usepackage{graphicx}

\title{ThaiOCRBench: A Task-Diverse Benchmark for Vision-Language Understanding in Thai}

\author{
  \textbf{Surapon~Nonesung},
  \textbf{Teetouch~Jaknamon},
  \textbf{Sirinya~Chaiophat}, \\
  \textbf{Natapong~Nitarach},
  \textbf{Chanakan~Wittayasakpan},
  \textbf{Warit~Sirichotedumrong},\\
  \textbf{Adisai~Na-Thalang},
  \textbf{Kunat~Pipatanakul}\\[1ex]
  SCB 10X R\&D,\ SCB 10X,\ SCBX Group,\ Thailand\\[0.5ex]
  \small{
    \href{mailto:surapon@scb10x.com}{surapon@scb10x.com},
    \href{mailto:kunat@scb10x.com}{kunat@scb10x.com}
  }
}

\begin{document}
\maketitle
\begin{abstract}
We present ThaiOCRBench, the first comprehensive benchmark for evaluating vision-language models (VLMs) on Thai text-rich visual understanding tasks. Despite recent progress in multimodal modeling, existing benchmarks predominantly focus on high-resource languages, leaving Thai underrepresented, especially in tasks requiring document structure understanding. ThaiOCRBench addresses this gap by offering a diverse, human-annotated dataset comprising 2,808 samples across 13 task categories. We evaluate a wide range of state-of-the-art VLMs in a zero-shot setting, spanning both proprietary and open-source systems. Results show a significant performance gap, with proprietary models (e.g., Gemini 2.5 Pro) outperforming open-source counterparts. Notably, fine-grained text recognition and handwritten content extraction exhibit the steepest performance drops among open-source models. Through detailed error analysis, we identify key challenges such as language bias, structural mismatch, and hallucinated content. ThaiOCRBench provides a standardized framework for assessing VLMs in low-resource, script-complex settings, and provides actionable insights for improving Thai-language document understanding.
\end{abstract}

\section{Introduction}

Vision-Language Models (VLMs) have demonstrated strong performance across a variety of multimodal tasks, including image captioning, visual question answering (VQA), and visual grounding. These advancements are primarily driven by transformer-based architectures and large-scale pretraining on image–text pairs. However, despite these gains, VLMs continue to face significant challenges when processing text-heavy images, particularly documents characterized by complex layouts, dense text, and multilingual scripts~\citep{hu-etal-2024-mplug, zhang2025documentparsingunveiledtechniques}. These limitations are especially evident in low-resource languages like Thai, where both linguistic and structural characteristics are underrepresented in current training corpora.

Most existing VLMs are trained and evaluated on English-centric datasets that fail to capture the unique features of Thai, such as the absence of inter-word spacing, the presence of stacked diacritics, and the diversity of document formats. While some multilingual VLMs nominally support Thai at the tokenization and inference levels, their performance on Thai-specific tasks has not been systematically assessed. Furthermore, the lack of standardized, human-annotated benchmarks for Thai text-rich vision tasks hinders rigorous evaluation and slows progress toward developing robust, language-inclusive VLMs.

In contrast, numerous benchmarks have been developed for high-resource languages, especially English. Early efforts focused on scene text recognition (e.g., IIIT5K~\citep{MishraBMVC12}, SVT~\citep{WangICCV11}), followed by more complex datasets such as TextVQA~\citep{singh2019towards}, DocVQA~\citep{mathew2021docvqadatasetvqadocument}, and ChartQA~\citep{masry-etal-2022-chartqa}. Additional benchmarks such as FUNSD~\citep{jaume2019funsd} and SROIE~\citep{huang2019icdar2019} target structured document understanding through Key information extraction. In the Thai context, existing datasets such as the NECTEC Thai OCR corpus, BEST2019~\citep{nectec2020best2019}, and Burapha-TH~\citep{app12084083} primarily support low-level tasks such as character or handwritten extraction, offering limited coverage of higher-level reasoning. Some small-scale efforts address scene text~\citep{suwanwiwat2021multiscript}, but comprehensive benchmarks for tasks such as layout parsing, relation extraction, or document-level VQA remain unavailable.

Recent evaluation frameworks OCRBench~\citep{Liu_2024}, OCRBench v2~\citep{fu2024ocrbenchv2improvedbenchmark}, and CC-OCR~\citep{yang2024ccocr} cover a broad set of tasks across document understanding and visual reasoning. However, these benchmarks overwhelmingly focus on high-resource languages, with Thai either underrepresented or excluded. Although recent multilingual efforts such as MTVQA~\citep{tang2024mtvqa} and PM4Bench~\citep{gao2025pm4benchparallelmultilingualmultimodal} include Thai, they are limited in task diversity and primarily address basic VQA.

To address this gap, we propose \textbf{ThaiOCRBench}, the first comprehensive benchmark designed to evaluate VLMs on Thai language, text-rich visual tasks. ThaiOCRBench contains 2,808 human-annotated samples spanning 13 task categories and diverse domains, including 
Chart parsing, Table parsing, Document parsing, Fine-grained text recognition, Full-page OCR, Handwritten content extraction, Text recognition, Key information extraction, Key information mapping, Document classification, Diagram VQA, Cognition VQA, and Infographics VQA.


This benchmark enables a focused investigation of the following research questions:
\begin{itemize}
\item \textbf{RQ1}: \textit{How well do current VLMs generalize to Thai-language text-rich visual tasks?}
\item \textbf{RQ2}: \textit{What are the common failure modes of open-source VLMs on these tasks, and how do they vary across tasks and model scales?}
\end{itemize}

\begin{figure*}
\centering
\includegraphics[width=0.7\textwidth]{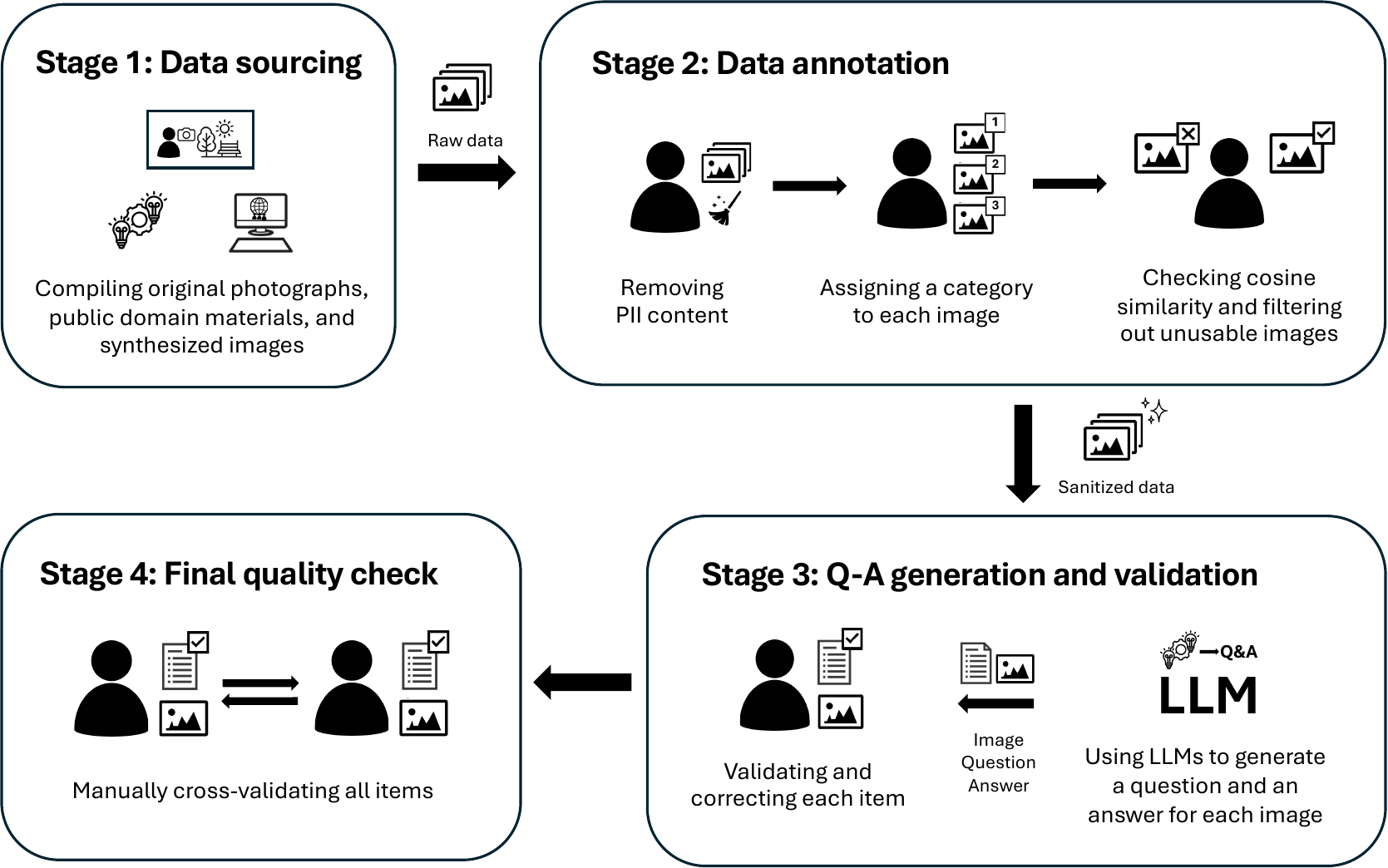}
\caption{Overview of the ThaiOCRBench data collection and annotation pipeline.}
\label{fig:data_flow}
\end{figure*}

To explore these questions, we conduct two complementary studies. For \textbf{RQ1}, we perform a systematic zero-shot evaluation of both proprietary and open-source VLMs on ThaiOCRBench. For \textbf{RQ2}, we carry out a qualitative error analysis of open-source models to identify prevalent failure modes and characterize performance gaps.

Our findings indicate that proprietary models particularly Gemini 2.5 Pro~\citep{comanici2025gemini25pushingfrontier} consistently outperform open-source counterparts. Among open-source models, Qwen2.5-VL 72B~\citep{bai2025qwen25vltechnicalreport} achieves the highest overall performance, though a notable gap remains. Detailed analysis reveals three dominant error patterns in open-source models: (1) Language Bias and Code-switching, (2) Structural Mismatch, and (3) Incorrect content.

\textbf{Contributions.} Our work makes the following key contributions:
\begin{itemize}
\item We introduce \textbf{ThaiOCRBench}, the first multi-task benchmark tailored for Thai-language vision-language understanding, with 2,808 human-annotated samples covering 13 task types. The dataset\footnote{https://huggingface.co/datasets/scb10x/ThaiOCRBench} and evaluation code\footnote{https://github.com/scb-10x/ThaiOCRBench} are publicly available to facilitate future research and reproducibility.
\item We establish zero-shot baselines for state-of-the-art VLMs, spanning both proprietary and open-source systems, enabling standardized evaluation for Thai-language document tasks.
\item We conduct an error analysis of open-source models, highlighting common limitations and offering insights for future improvements in Thai-specific VLM capabilities.
\end{itemize}
\section{Related Work}

\subsection{Vision-Language Models with Thai Support}

Most vision-language models (VLMs) have been developed and benchmarked primarily on high-resource languages, particularly English and Chinese. Recent advancements include both open-source models such as Gemma3~\citep{gemmateam2025gemma3technicalreport}, Qwen2.5-VL, and LLaMA3.2 Vision~\citep{llama3modelcard} and proprietary systems such as GPT-4o~\citep{openai2024gpt4technicalreport}, Gemini 2.5 Pro, and Claude Sonnet 4~\citep{claude2025}. These models demonstrate strong performance across various document understanding tasks and generally support multiple languages at the tokenization and inference levels.

However, their evaluations are typically restricted to multilingual benchmarks such as MTVQA, which primarily emphasize high-level tasks such as visual question answering (VQA). Systematic assessments of these models on Thai-specific tasks, especially those requiring fine-grained reasoning over structured and complex content such as tables, forms, and charts remain limited. Consequently, the extent to which current VLMs generalize to Thai-language, text-rich scenarios is still largely unexplored. This work addresses this gap by introducing a benchmark specifically designed to enable systematic evaluation of VLMs across a wide range of Thai-language vision tasks.

\subsection{Benchmarks for Thai Text-Rich Vision Tasks}

Benchmark resources for Thai-language vision tasks remain limited in both task diversity and complexity. Existing datasets focus predominantly on low-level recognition. For instance, the NECTEC Thai OCR corpus provides printed Thai text images for character-level optical character recognition (OCR), while BEST2019 offers annotated handwritten lines for offline handwritten extraction. Similarly, the Burapha-TH dataset targets isolated character and syllable recognition.

While these datasets are valuable for developing foundational OCR systems, they lack the structural and semantic annotations necessary to support higher-level tasks such as element parsing, relation extraction, or VQA. Moreover, no existing benchmark integrates a diverse set of Thai-language vision tasks within a unified framework. This limits comprehensive evaluation of models in realistic, document-centric scenarios.


\begin{figure*}[t]
\centering
\begin{subfigure}[t]{0.49\textwidth}
\centering
\includegraphics[width=\textwidth]{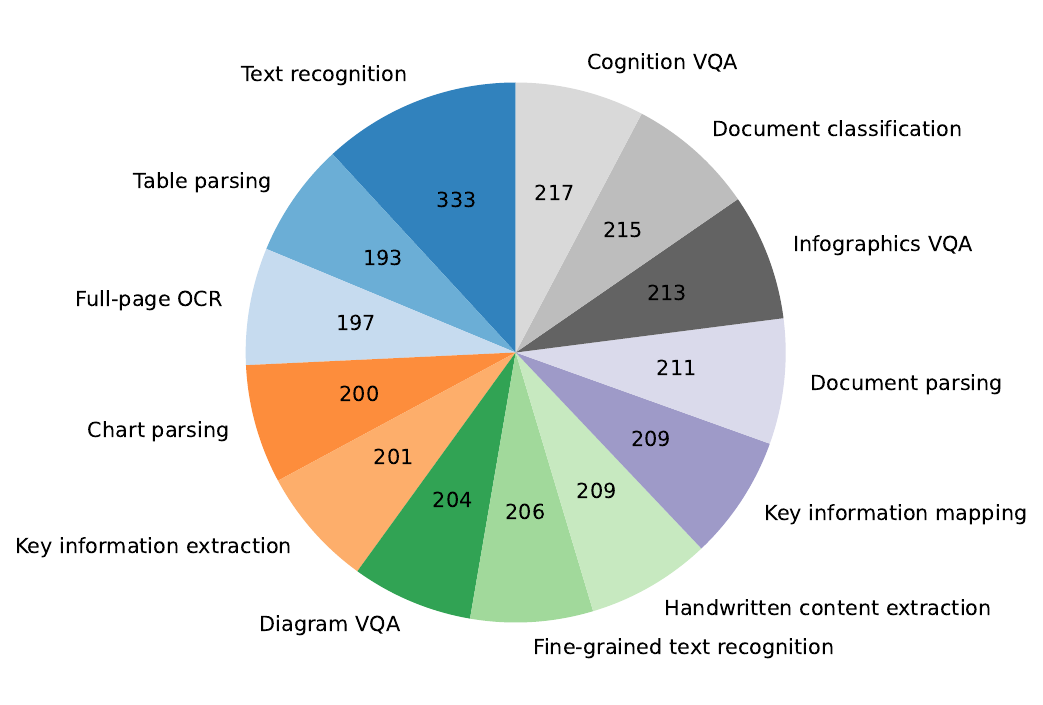}
\caption{Distribution of the 13 task types in ThaiOCRBench.}
\label{fig:task_type_distribution}
\end{subfigure}
\hfill
\begin{subfigure}[t]{0.49\textwidth}
\centering
\includegraphics[width=\textwidth]{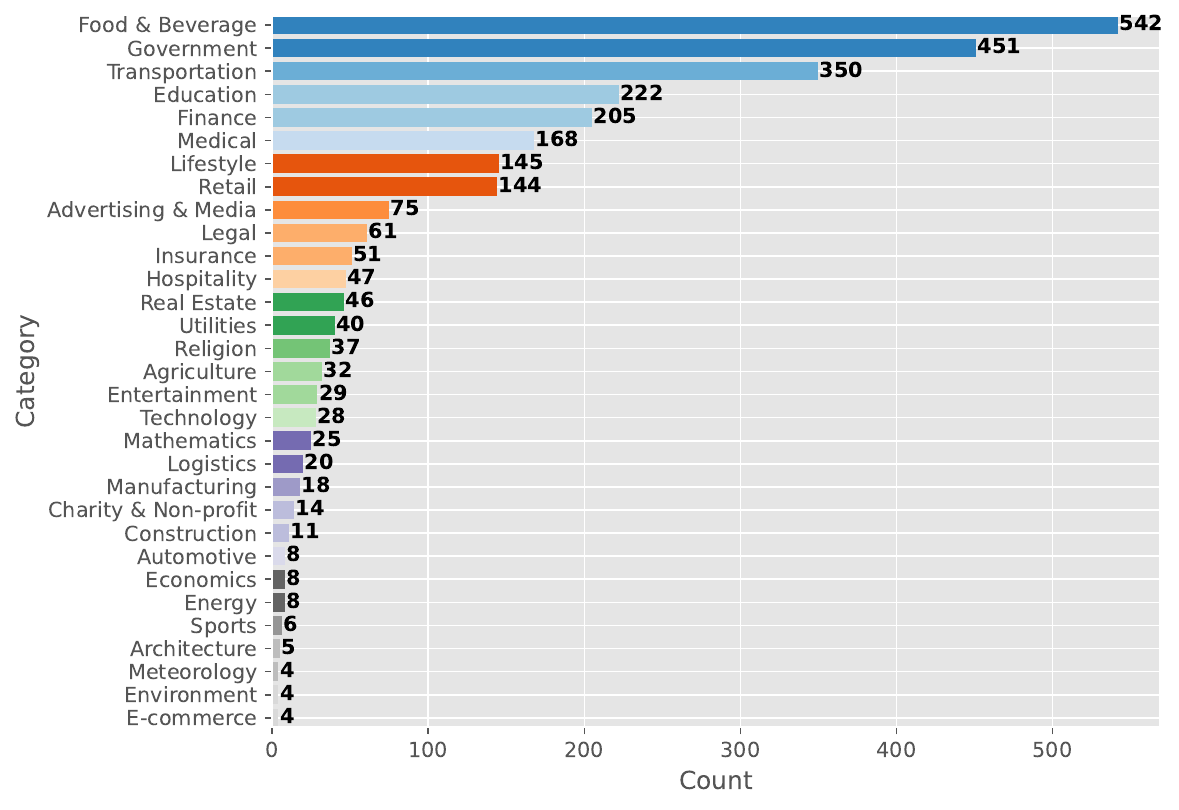}
\caption{Domain-specific category distribution.}
\label{fig:domain_category_distribution}
\end{subfigure}
\caption{Distributional statistics of ThaiOCRBench, illustrating the coverage across task types and domain-specific categories.}
\label{fig:thaiocrbench_distribution}
\end{figure*}

\section{ThaiOCRBench: Dataset Construction}
The construction of ThaiOCRBench was a multi-stage process guided by clear design principles to ensure its cultural relevance, diversity, and overall quality. This section details these principles, the task definitions, the data sourcing and annotation pipeline, and the final dataset statistics.

\subsection{Design Principles}

The construction of ThaiOCRBench was guided by two core design principles: cultural specificity and data diversity.

\textbf{Cultural specificity} emphasizes the inclusion of content that reflects linguistic, visual, and contextual elements unique to Thai cultures. This ensures that the benchmark evaluates model performance in authentically Thai scenarios rather than relying on generalized or translated content. Examples include visual elements requiring local cultural knowledge, such as Bangkok’s color-coded public transportation signage, and culturally specific symbols, such as prohibition signs against durians in public areas. Additionally, the dataset incorporates linguistically complex content, such as Pali-Sanskrit chants written in Thai script, which feature rare characters and vocabulary that are typically absent from standard web-based corpora.

\textbf{Data diversity} ensures broad representation across domains, text modalities, and visual styles. The benchmark includes a variety of document types (e.g., government reports, restaurant menus, medical forms), text formats (e.g., machine-printed, handwritten, poetic verse), and typographic styles. This also includes both traditional Thai "headed" scripts and modern "headless" variants. The latter introduces significant recognition challenges due to their visual similarity to Latin characters. For example, the headless form of the Thai letter "Nor Nu" closely resembles the lowercase Latin letter "u", creating substantial ambiguity for OCR systems.

Representative examples illustrating these principles are provided in
Appendix~\ref{appendix:dataset_samples}, highlighting the benchmark’s emphasis on linguistic complexity and real-world variability.



\subsection{Task Categories Definition}

ThaiOCRBench is a multi-task benchmark comprising 13 task types designed to evaluate the capabilities of vision-language models (VLMs) in processing Thai text-rich visual content. While OCRBench v2 provides a broader task set aggregated from multiple datasets, we adopt it as a reference due to its comprehensive coverage and structured evaluation methodology.

In contrast to OCRBench v2, ThaiOCRBench emphasizes a focused set of linguistically and structurally challenging tasks tailored specifically to the Thai language. All images in the dataset are newly collected and manually annotated to reflect authentic layouts, localized formats, and language-specific phenomena. The task categories are defined in Appendix~\ref{appendix:task_cat_def}

\subsection{Data Sourcing and Annotation}
As illustrated in Figure~\ref{fig:data_flow}, the dataset was constructed through a four-stage pipeline designed to ensure data diversity, ethical compliance, and annotation quality.


\textbf{Stage 1: Data Sourcing.}
Images were collected from a variety of sources, including original photographs taken in public spaces, publicly available materials, and licensed commercial datasets. For sensitive document types such as identification cards and legal certificates, synthetic documents were programmatically generated to avoid privacy concerns. All images underwent a sanitization process in which human annotators manually removed or obscured personally identifiable information (PII), such as faces, names, and identification numbers.

\textbf{Stage 2: Data Annotation.}
Human annotators categorized each image based on content type and assigned relevant metadata, including source information, licensing details, and descriptive tags. To ensure data uniqueness and reduce redundancy, pairwise cosine similarity was computed across image embeddings. Instances of high similarity such as images with near-identical angles, fonts, or layouts in the same task category were flagged and reviewed, and duplicates or near-duplicates were removed accordingly. Detailed annotation guidelines are provided in Appendix~\ref{appendix:annotator_guideline}.

\textbf{Stage 3: Question–Answer Generation and Validation.}
We employed multiple large language models (LLMs), including GPT-4o, Gemini 2.5 Pro, and Azure AI Services~\citep{azure_document_intelligence}, to generate initial question–answer (QA) pairs for each image. Human annotators then reviewed these outputs, selecting or refining the most suitable pairs based on task-specific guidelines. As many generated QA pairs were found to be inaccurate or misaligned with the visual content, substantial manual revision or rewriting was conducted to ensure correctness and task relevance.


\textbf{Stage 4: Final Quality Control.}
A separate team of annotators conducted a final review of all dataset entries. Each item comprising the image, associated question, and answer was assessed for coherence, accuracy, and alignment with the intended task definitions. Only items that met all quality standards were retained in the final benchmark.

\begin{table*}[t]
    \centering
    \scriptsize
    \begin{tabular}{lccccccccccccc|c}
        \toprule
    \textbf{Model} &
    \multicolumn{3}{c}{\textbf{TED}} &
    \multicolumn{4}{c}{\textbf{BMFL}} &
    \multicolumn{2}{c}{\textbf{F1}} &
    \multicolumn{4}{c}{\textbf{ANLS}} &
    \rotatebox{90}{} \\
    \cmidrule(lr){2-4}
    \cmidrule(lr){5-8}
    \cmidrule(lr){9-10}
    \cmidrule(lr){11-14}
    & \rotatebox{90}{Chart parsing} &
        \rotatebox{90}{Table parsing} &
        \rotatebox{90}{Doc. parsing} &
        \rotatebox{90}{Fine-grained Rec.} &
        \rotatebox{90}{Full-page OCR} &
        \rotatebox{90}{Handwritten} &
        \rotatebox{90}{Text recognition} &
        \rotatebox{90}{Info. extraction} &
        \rotatebox{90}{Info. mapping} &
        \rotatebox{90}{Doc. classification} &
        \rotatebox{90}{Diagram VQA} &
        \rotatebox{90}{Cognition VQA} &
        \rotatebox{90}{Infographics VQA} &
        \rotatebox{90}{Average score} \\
        \midrule
        \rowcolor{Gray}
        \multicolumn{15}{l}{\textbf{OCR-specialized model}} \\
        EasyOCR        & - & - & - & - & 0.61 & 0.124 & 0.458 & - & - & - & - & - & - & - \\
Tesseract OCR & - & - & - & - & 0.614 & 0.071 & 0.271 & - & - & - & - & - & - & - \\
        \rowcolor{Gray}
        \multicolumn{15}{l}{\textbf{Proprietary model}} \\
Gemini 2.5 Pro        & 0.812 & \textbf{0.686} & \textbf{0.587} &\textbf{0.499} & \textbf{0.897} & \textbf{0.714} & \textbf{0.910} & \textbf{0.658} & \textbf{0.863} & 0.943 & \textbf{0.766} & \textbf{0.872} & \textbf{0.898} & \textbf{0.777} \\
Claude Sonnet 4              & \textbf{0.817} & 0.650 & 0.543 & 0.214 & 0.661 & 0.301 & 0.686 & 0.452 & 0.675 & 0.879 & 0.379 & 0.657 & 0.613 & 0.579 \\
GPT-4o              & 0.766 & 0.571 & 0.515 & 0.254 & 0.610 & 0.489 & 0.778 & 0.546 & 0.734 & \textbf{0.973} & 0.562 & 0.796 & 0.791 & 0.645 \\
\rowcolor{Gray}
        \multicolumn{15}{l}{\textbf{Open-source model}} \\
Gemma3 27B          & 0.783 & 0.519 & 0.350 & 0.144 & 0.608 & 0.280 & 0.561 & 0.389 & 0.574 & 0.831 & 0.309 & 0.514 & 0.552 & 0.493 \\
Gemma3 12B          & 0.704 & 0.395 & 0.358 & 0.084 & 0.504 & 0.225 & 0.433 & 0.300 & 0.558 & 0.770 & 0.270 & 0.428 & 0.471 & 0.423 \\
Gemma3 4B           & 0.635 & 0.322 & 0.355 & 0.089 & 0.363 & 0.143 & 0.233 & 0.225 & 0.493 & 0.683 & 0.129 & 0.349 & 0.343 & 0.336 \\
Qwen2.5-VL 72B         & \underline{0.801} & \underline{0.549} & \underline{0.454} & 0.147 & \underline{0.720} & \underline{0.393} & \underline{0.749} & \underline{0.497} & \underline{0.719} & \underline{0.914} & \underline{0.519} & \underline{0.746} & \underline{0.782} & \underline{0.615} \\
Qwen2.5-VL 32B         & 0.765 & 0.483 & 0.334 & 0.139 & 0.553 & 0.280 & 0.635 & 0.394 & 0.708 & 0.860 & 0.409 & 0.650 & 0.681 & 0.530 \\
Qwen2.5-VL 7B          & 0.712 & 0.509 & 0.308 & \underline{0.218} & 0.631 & 0.314 & 0.597 & 0.354 & 0.623 & 0.862 & 0.416 & 0.702 & 0.763 & 0.539 \\
Qwen2.5-VL 3B          & 0.650 & 0.431 & 0.338 & 0.130 & 0.430 & 0.210 & 0.475 & 0.284 & 0.481 & 0.821 & 0.308 & 0.532 & 0.550 & 0.434 \\
InternVL3 78B       & 0.768 & 0.440 & 0.434 & 0.073 & 0.167 & 0.158 & 0.069 & 0.300 & 0.572 & 0.759 & 0.217 & 0.306 & 0.367 & 0.356 \\
InternVL3 14B       & 0.760 & 0.399 & 0.405 & 0.059 & 0.184 & 0.140 & 0.038 & 0.334 & 0.534 & 0.712 & 0.170 & 0.321 & 0.352 & 0.339 \\
InternVL3 8B        & 0.731 & 0.423 & 0.298 & 0.052 & 0.157 & 0.127 & 0.033 & 0.252 & 0.480 & 0.698 & 0.154 & 0.269 & 0.305 & 0.306 \\
Aya-Vision 8B       & 0.567 & 0.229 & 0.322 & 0.027 & 0.080 & 0.075 & 0.005 & 0.056 & 0.187 & 0.466 & 0.058 & 0.115 & 0.123 & 0.178 \\
Kimi-VL-A3B-Instruct& 0.404 & 0.373 & 0.327 & 0.026 & 0.105 & 0.091 & 0.013 & 0.176 & 0.159 & 0.551 & 0.113 & 0.189 & 0.261 & 0.214 \\
SmolVLM2 2.2B        & 0.015 & 0.042 & 0.134 & 0.030 & 0.049 & 0.048 & 0.000 & 0.003 & 0.000 & 0.135 & 0.010 & 0.017 & 0.030 & 0.039 \\
Pixtral 12B           & 0.637 & 0.380 & 0.334 & 0.039 & 0.113 & 0.091 & 0.018 & 0.154 & 0.393 & 0.671 & 0.094 & 0.191 & 0.270 & 0.260 \\
Phi-3 vision 4B              & 0.475 & 0.186 & 0.202 & 0.034 & 0.039 & 0.057 & 0.006 & 0.119 & 0.209 & 0.269 & 0.039 & 0.142 & 0.148 & 0.148 \\
Skywork-R1V-38B     & 0.756 & 0.418 & 0.385 & 0.074 & 0.181 & 0.128 & 0.055 & 0.344 & 0.558 & 0.765 & 0.136 & 0.256 & 0.304 & 0.335 \\
Phi-4 multimodal 5B               & 0.591 & 0.212 & 0.237 & 0.028 & 0.050 & 0.063 & 0.003 & 0.065 & 0.237 & 0.316 & 0.038 & 0.129 & 0.131 & 0.162 \\
Llama 3.2-Vision 11B           & 0.222 & 0.326 & 0.252 & 0.051 & 0.207 & 0.145 & 0.237 & 0.097 & 0.485 & 0.769 & 0.163 & 0.368 & 0.424 & 0.288 \\
MiniCPM-o 2.6 8B           & 0.497 & 0.181 & 0.170 & 0.046 & 0.082 & 0.075 & 0.008 & 0.050 & 0.256 & 0.628 & 0.106 & 0.206 & 0.241 & 0.196 \\
        \bottomrule
    \end{tabular}
    \caption{Performance comparison of proprietary and open-source models on ThaiOCRBench. Tasks are grouped by evaluation metric: \textbf{TED} (Chart, Table, Doc Parsing), \textbf{BMFL} (Generation and Recognition tasks), \textbf{F1} (Information extraction tasks), and \textbf{ANLS} (Understanding/VQA tasks). Bold values denote the best proprietary model; underlined values denote the best open-source model.}
    \label{tab:doc_task_model_eval}
\end{table*}


\subsection{Dataset Statistics}
ThaiOCRBench consists of 2,808 images with human-annotated question-answer pairs. Figure~\ref{fig:task_type_distribution} illustrates the distribution across the 13 task types, while Figure~\ref{fig:domain_category_distribution} shows the coverage of domain-specific categories. Token length statistics for questions and answers are provided in Appendix~\ref{fig:qa_distribution}.

\section{Experimental Design}
We design and conduct two complementary studies using the ThaiOCRBench benchmark. To address \textbf{RQ1}, we perform zero-shot evaluations of state-of-the-art vision-language models (VLMs), encompassing both proprietary and open-source systems, to assess their effectiveness on Thai language, text-rich visual tasks (Section~\ref{rq1_eval}). To address \textbf{RQ2}, we conduct a qualitative error analysis of open-source models, categorizing failure cases into three primary types. This analysis aims to identify key limitations and inform strategies for narrowing the performance gap relative to proprietary models (Section~\ref{rq2}).

\subsection{Evaluation Models}
We evaluate a range of state-of-the-art vision-language models (VLMs), encompassing both proprietary and open-source systems, to assess their zero-shot performance on ThaiOCRBench. The selected models vary in architectural design, training data, and parameter scale, allowing for a comprehensive comparison across model families.

All evaluations are conducted in a zero-shot setting using the \texttt{vLLM} inference engine \footnote{\url{https://github.com/vllm-project/vllm}} with greedy decoding. The models assessed include proprietary systems such as Gemini 2.5 Pro, Claude Sonnet 4, and GPT-4o , as well as open-source models including Qwen2.5-VL, Gemma3, LLaMA3.2 Vision, InternVL 3~\citep{zhu2025internvl3exploringadvancedtraining}, Aya-Vision~\citep{dash2025ayavisionadvancingfrontier}, Kimi-VL~\citep{kimiteam2025kimivltechnicalreport}, SmolVLM~\citep{marafioti2025smolvlmredefiningsmallefficient}, Pixtral~\citep{agrawal2024pixtral12b}, Phi-3~\citep{abdin2024phi3technicalreporthighly}, Phi-4~\citep{abdin2024phi4technicalreport}, Skywork-R1V~\citep{peng2025skyworkr1vpioneeringmultimodal} and MiniCPM-o 2.6~\citep{yao2024minicpm}.

To provide additional context and bridge evaluation with conventional OCR pipelines, we also include OCR-specialized baselines—Tesseract OCR~\citep{smith2007overview} and EasyOCR~\citep{easyocr}—as reference systems. Although these models are not designed for multimodal reasoning, they serve as useful lower-bound baselines for text extraction accuracy in Thai-language documents.

\subsection{Evaluation Metrics}

We adopt evaluation metrics from OCRBench v2.

\textbf{Structural Understanding Tasks.}
For tasks involving the reconstruction of document layout and hierarchical content such as \textit{Table parsing}, \textit{Chart parsing}, and \textit{Document parsing}, we employ the Tree Edit Distance (TED) metric~\citep{zhong2020imagebasedtablerecognitiondata}, which quantifies structural similarity between predicted and reference outputs. TED is particularly suited to evaluating nested or hierarchical formats where layout consistency is critical.

\textbf{Text Generation and Recognition Tasks.}
For tasks requiring the transcription or generation of text such as \textit{Fine-grained text recognition}, \textit{Full-page OCR}, and \textit{Handwritten content extraction}, we report multiple complementary metrics to assess both character-level accuracy and linguistic fidelity. These include BLEU~\citep{papineni-etal-2002-bleu}, METEOR~\citep{banerjee-lavie-2005-meteor}, F1-score, and Normalized Levenshtein Similarity (NLS)~\citep{biten2019scenetextvisualquestion}. We average these into a single composite metric, referred to as BMFL.


\textbf{Structured Prediction Tasks.}
For \textit{Key information extraction} and \textit{Key information mapping}, we use the F1-score to evaluate precision and recall in entity-level prediction. This metric is appropriate for scenarios where exact field alignment is required and partial matches are penalized.

\textbf{Textual Understanding and Question Answering Tasks.}
For tasks involving semantic understanding and short-form textual generation such as \textit{Text recognition}, \textit{Document classification}, \textit{Diagram VQA}, \textit{Cognition VQA}, and \textit{Infographics VQA}, we adopt the Average Normalized Levenshtein Similarity (ANLS)~\citep{biten2019scenetextvisualquestion}. ANLS measures similarity by normalizing the edit distance between predicted and reference responses, allowing for partial credit when predictions are close but not exact.

Although each metric is calculated differently, all are designed such that higher scores indicate better performance. To facilitate model comparison across diverse tasks, we also report the average score as an overall performance indicator.

\section{Experiment Results}
\subsection{Zero-Shot Performance Evaluation}
\label{rq1_eval}

To address \textbf{RQ1}, we evaluate state-of-the-art vision-language models (VLMs) on ThaiOCRBench under a zero-shot greedy decoding setting.

\begin{figure*}[t]
    \centering
    \begin{subfigure}[t]{0.49\textwidth}
        \centering
        \includegraphics[width=\textwidth]{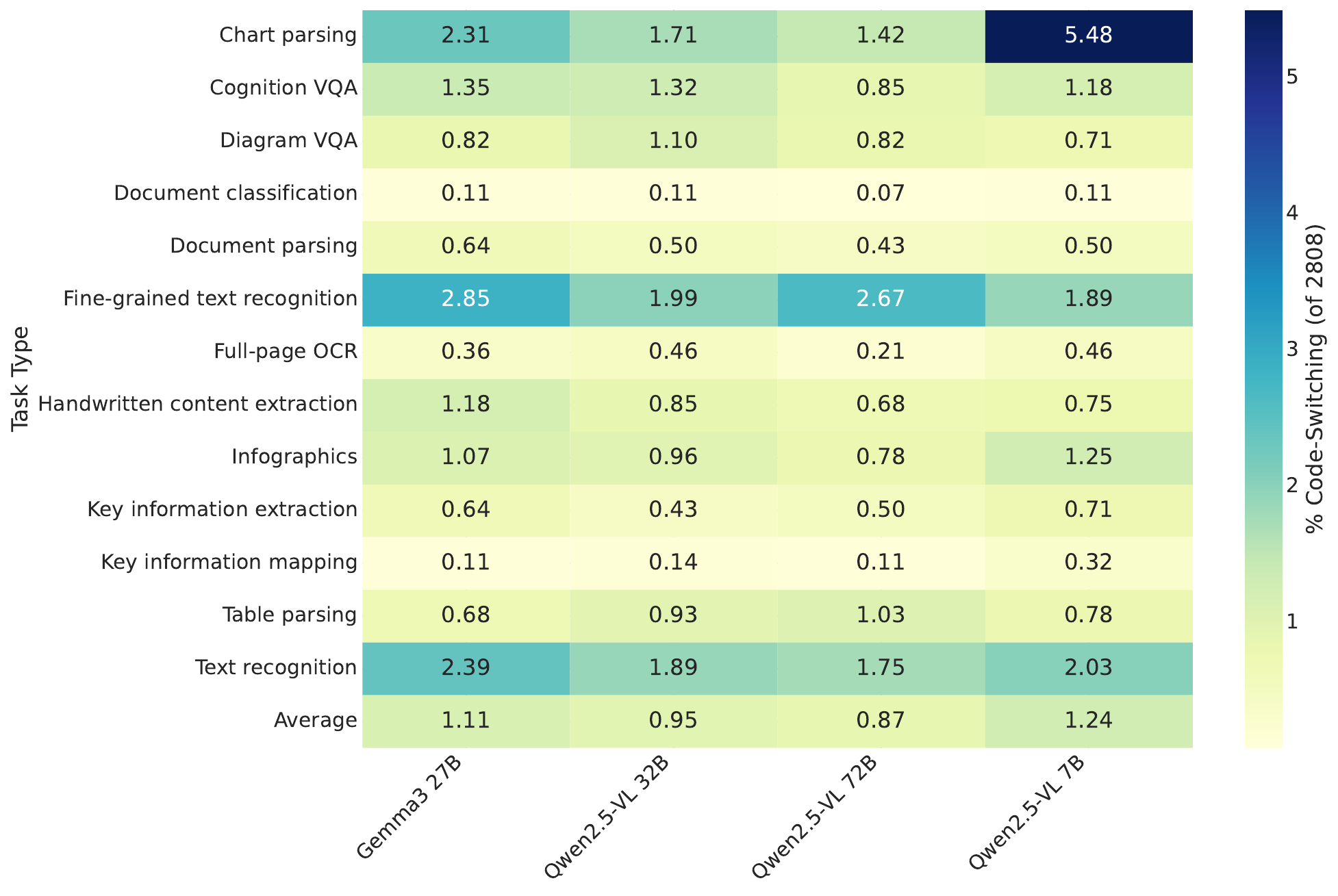}
        \caption{Language Bias and Code-Switching behavior across models, measured by the proportion of data samples exhibiting language switching, normalized by the total number of samples.}
        \label{fig:model_comparison_error}
    \end{subfigure}
    \hfill
    \begin{subfigure}[t]{0.49\textwidth}
        \centering
        \includegraphics[width=\textwidth]{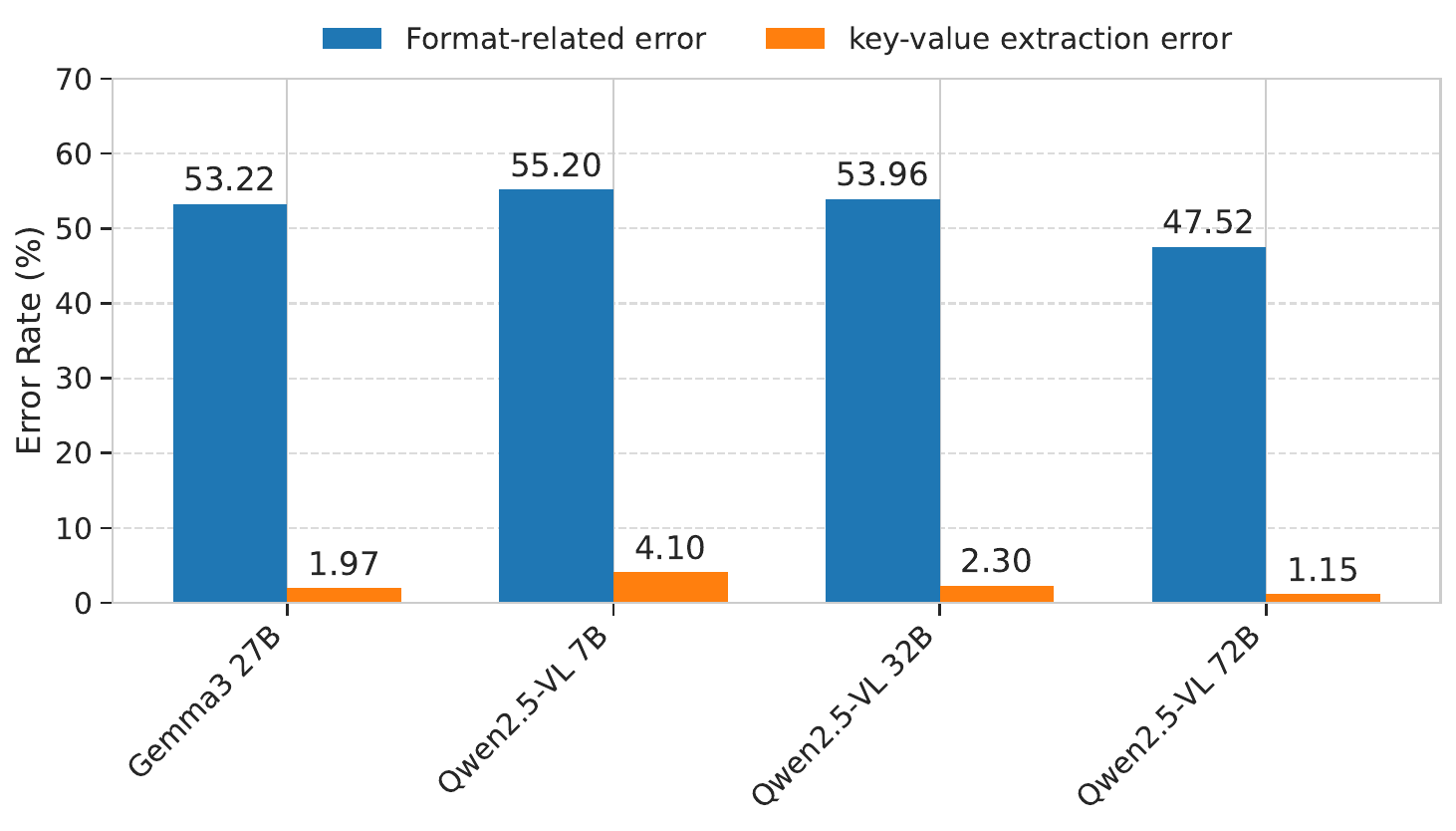}
        \caption{Structural Mismatch across models, measured by the proportion of data points exhibiting structural inconsistencies, normalized by the total number of error instances.}
        \label{fig:task_error_heatmap}
    \end{subfigure}
    \caption{Qualitative error breakdown across Top-4 models.}
    \label{fig:error_type_distribution}
\end{figure*}

\begin{table*}[t]
    \centering
    \tiny
    \renewcommand{\arraystretch}{1.2}
    \begin{adjustbox}{max width=\textwidth}
    \begin{tabular}{lcccccccccccccccc}
        \toprule
         & \multicolumn{4}{c}{Substitutions ($\times10^4$)} 
         & \multicolumn{4}{c}{Deletions ($\times10^4$)} 
         & \multicolumn{4}{c}{Insertions ($\times10^4$)} 
         & \multicolumn{4}{c}{Correct ($\times10^4$)} \\
         \cmidrule(r){2-5} \cmidrule(r){6-9} \cmidrule(r){10-13} \cmidrule(r){14-17}
         Task & G27 & Q32 & Q72 & Q7 & G27 & Q32 & Q72 & Q7 & G27 & Q32 & Q72 & Q7 & G27 & Q32 & Q72 & Q7 \\
        \midrule
        Fine-grained text       & 0.23 & 0.27 & 0.18 & 0.19 & 0.16 & 0.11 & 0.24 & 0.17 & 0.24 & 1.76 & 1.12 & 2.09 & 0.14 & 0.16 & 0.12 & 0.19 \\
        Infographics VQA           & 0.19 & 0.15 & 0.07 & 0.10 & 0.30 & 0.15 & 0.25 & 0.25 & 0.31 & 0.44 & 0.08 & 0.12 & 0.53 & 0.73 & 0.70 & 0.67 \\
        Chart parsing           & 4.89 & 3.33 & 3.34 & 3.72 & 3.49 & 3.42 & 2.69 & 3.87 & 5.44 & 7.00 & 6.92 &16.96 & 9.10 &10.74 &11.45 & 9.89 \\
        Cognition VQA           & 0.12 & 0.09 & 0.06 & 0.09 & 0.21 & 0.17 & 0.19 & 0.17 & 0.29 & 0.13 & 0.05 & 1.62 & 0.45 & 0.52 & 0.54 & 0.53 \\
        Diagram VQA             & 0.15 & 0.15 & 0.10 & 0.12 & 0.22 & 0.16 & 0.23 & 0.22 & 0.20 & 5.14 & 0.07 & 0.22 & 0.23 & 0.30 & 0.27 & 0.26 \\
        Document parsing        & 7.64 &11.74 & 6.02 & 9.02 &10.05 & 5.09 & 5.33 & 8.47 &\textbf{56.76} &\textbf{41.24} &\textbf{98.52} &\textbf{38.98} &18.61 &19.46 &24.94 &18.80 \\
        Full-page OCR           & 4.69 & 6.25 & 2.97 & 3.44 & 6.07 & 2.47 & 2.25 & 4.51 & 5.55 &59.66 & 8.85 &21.65 &17.87 &19.91 &23.40 &20.67 \\
        Table parsing           & 6.46 & 7.84 & 5.46 & 6.05 &\textbf{18.38} &\textbf{14.64} &\textbf{15.97} &\textbf{15.37} & 6.32 &30.04 &12.71 &18.07 &28.17 &30.54 &31.60 &31.60 \\
        Text recognition        & 0.94 & 0.79 & 0.45 & 0.51 & 0.88 & 0.54 & 0.55 & 1.66 & 2.21 & 7.91 & 2.88 & 7.51 & 4.73 & 5.21 & 5.55 & 4.37 \\
        Handwritten             & 0.87 & 0.97 & 0.72 & 0.73 & 0.48 & 0.25 & 0.49 & 0.63 &19.53 &12.39 & 3.18 &10.20 & 1.23 & 1.36 & 1.37 & 1.22 \\
        Key info. extraction          & 0.49 & 0.56 & 0.34 & 0.69 & 1.22 & 0.68 & 0.73 & 1.12 & 1.41 & 6.25 & 7.53 & 6.18 & 6.33 & 6.79 & 6.96 & 6.22 \\
        Key info. mapping             & 0.56 & 0.51 & 0.53 & 0.55 & 0.91 & 0.89 & 0.91 & 0.89 & 0.37 & 0.48 & 0.30 & 1.98 & 4.02 & 4.11 & 4.06 & 4.06 \\
        Doc. classification          & 0.03 & 0.02 & 0.01 & 0.02 & 0.01 & 0.00 & 0.00 & 0.00 & 0.03 & 0.04 & 0.02 & 0.06 & 0.24 & 0.25 & 0.26 & 0.25 \\
        \midrule
        \textbf{Total}          & 27.26 & 32.67 & 20.25 & 25.23 & 42.38 & 28.57 & 29.83 & 37.33 & 98.66 & 172.48 & 142.23 & 125.64 & 91.65 & 100.08 & 111.22 & 98.73 \\
        \bottomrule
    \end{tabular}
    \end{adjustbox}
    \caption{Incorrect Content analysis across models. Character-level errors are reported based on the Character Error Rate (CER), including substitutions, deletions, insertions, and correct tokens, aggregated across tasks for the top-4 models. All values are scaled by $10^4$ for readability. Model abbreviations: G27 = \textit{Gemma3 27B}, Q32 = \textit{Qwen2.5-VL 32B}, Q72 = \textit{Qwen2.5-VL 72B}, Q7 = \textit{Qwen2.5-VL 7B}.}

    \label{tab:error_components_by_model}
\end{table*}



\textbf{Results.}
Table~\ref{tab:doc_task_model_eval} summarizes the performance of proprietary and open-source models. Proprietary models consistently outperform open-source counterparts across most tasks. Gemini 2.5 Pro achieves the highest overall average score (0.777), ranking first in 11 out of 13 tasks. It performs particularly well in Key information mapping (0.863), Full-page OCR (0.897), and Text recognition (0.910).

GPT-4o also demonstrates strong performance, leading in Document classification (0.973) and Cognition VQA (0.796), with an overall score of 0.645. Claude Sonnet 4 follows with an average score of 0.579.

Among open-source models, Qwen2.5-VL 72B achieves the highest average (0.615), closing the gap with proprietary systems in tasks such as Document classification (0.914), Full-page OCR (0.720), and Cognition VQA (0.615). Other Qwen variants (32B and 7B) also perform competitively across multiple tasks.

\textbf{Discussion.} Our results reveal clear disparities in model performance across different ThaiOCRBench tasks, highlighting the varied demands of each subtask and the influence of evaluation metrics. For example, as shown in Table~\ref{tab:doc_task_model_eval}, models such as InternVL3 78B perform well on Chart parsing (0.768) but poorly on Text recognition (0.069). This discrepancy rises from differences in metric sensitivity. Chart parsing is evaluated using structure-aware metrics such as Tree Edit Distance (TED), which are less sensitive to minor text errors as long as the overall structure remains accurate. In contrast, Text recognition is assessed using normalized Levenshtein distance, a stricter metric that penalizes even small character-level mistakes, an especially challenging aspect when dealing with complex scripts such as Thai.


These results also reflect divergent model capabilities. InternVL3, for instance, may effectively capture document structure, benefiting layout-intensive tasks, but lacks robust Thai tokenization or pretraining, limiting its performance in script-level decoding.

We also observe that Document classification is comparatively less challenging. This task is framed as a constrained multiple-choice problem with seven predefined categories, reducing the need for complex reasoning. As a result, most models including smaller ones achieve strong performance. In contrast, Fine-grained text recognition remains the most difficult task, aligning with trends observed in English benchmarks such as OCRBench v2. It requires accurate localization and transcription of small text elements embedded in complex layouts capabilities that remain challenging even for large-scale models.

\subsection{Qualitative Error Analysis}
\label{rq2}

To address \textbf{RQ2}, we conduct a qualitative error analysis by categorizing common failure modes observed in model predictions. We focus on the top four open-source models based on their average performance on ThaiOCRBench (as reported in Table~\ref{tab:doc_task_model_eval}). Errors are classified into three primary categories: (1) \textit{Language Bias and Code-Switching}, (2) \textit{Structural Mismatch}, and (3) \textit{Incorrect Content}.

\textbf{Language Bias and Code-Switching.}
As shown in Figure~\ref{fig:model_comparison_error}, we analyze discrepancies between model predictions and ground-truth references using few-shot prompting with GPT-4o. Prompt details is in Appendix~\ref{appendix:code_switch_prompt}. This evaluation reveals two prevalent categories of linguistic errors: (i) \textit{language bias}, wherein models systematically default to non-Thai outputs despite receiving Thai inputs, and (ii) \textit{code-switching}, characterized by the inappropriate intermixing of Thai and non-Thai language elements within a single prediction. These phenomena underscore persistent challenges related to multilingual generalization and script fidelity in current model architectures.

\textbf{Structural Mismatch.} Structural errors fall into two categories, as illustrated in Figure~\ref{fig:task_error_heatmap}:
\begin{itemize}
\item \textit{Key–Value Extraction Errors}, found in tasks such as chart parsing, key information extraction, and key information mapping, where models fail to align fields with corresponding keys detected via rule-based validation.
\item \textit{Format-Related Errors}, common in document and table parsing, where predicted structural formats deviate from reference outputs (e.g., tag mismatches or missing components).
\end{itemize}

\textbf{Incorrect Content.}
To assess transcription fidelity, we calculate the Character Error Rate (CER), decomposed into standard edit operations: substitutions, deletions, insertions, and correct matches. This token-level analysis enables fine-grained assessment of recognition quality in tasks involving text transcription, particularly Full-page OCR and Fine-grained text recognition. The breakdown of these error components is provided in Table~\ref{tab:error_components_by_model}.

\textbf{Results.}
As illustrated in Figure~\ref{fig:model_comparison_error}, we observe a consistent trend in which larger models exhibit lower code-switching rates, indicating improved linguistic stability. In particular, Qwen2.5-VL 72B achieves the lowest average code-switching rate at 0.87\%, while Qwen2.5-VL 7B records the highest at 1.24\%, with a task-specific peak of 5.48\% in Chart parsing. These findings suggest that code-switching errors are more prevalent in tasks involving structured reasoning and complex visual layouts, such as tables and diagrams. Such content often contains multilingual or ambiguously formatted text, which appears to challenge smaller models with weaker cross-modal instruction following and limited multilingual grounding.

Figure~\ref{fig:task_error_heatmap} further highlights that format-related errors remain a major source of failure across all models, with error rates ranging from 47.52\% (Qwen2.5-VL 72B) to 55.20\% (Qwen2.5-VL 7B). These results underscore the persistent difficulty that VLMs face in layout-intensive tasks, particularly when structural fidelity is essential. In contrast, key–value extraction errors show greater sensitivity to the model scale, with error rates declining from 4.10\% in Qwen2.5-VL 7B to 1.15\% in Qwen2.5-VL 72B. This suggests that larger models are better able to capture semantic relationships and maintain accurate alignment between visual cues and entity representations.


As shown in Table~\ref{tab:error_components_by_model}, Gemma3 27B (G27) produces the highest overall deletion count (42.38), with particularly large deletion values in structurally demanding tasks such as Table Parsing (18.38) and Document Parsing (10.05). These results suggest that G27 is more prone to omission-related errors, reflecting a concise generation style.

In contrast, Qwen2.5-VL-32B (Q32) produces the highest number of insertions (172.48), driven largely by errors in Table parsing (30.04), Document parsing (41.24), and Full-page OCR (59.66). Other Qwen variants, such as Qwen2.5-VL 72B (Q72) and Qwen2.5-VL 7B (Q7), follow similar patterns, though with reduced insertion counts. In terms of correctly generated characters, Qwen2.5-VL 72B achieves the highest count (111.22), followed by Q32 (100.08), reflecting the models’ ability to produce a large amount of valid content despite higher insertion rates.

These trends reflect a core trade-off: Gemma favors conservative output, minimizing hallucinations, while Qwen prioritizes recall, risking more insertions.

\textbf{Discussion.} Our experimental results reveal clear trade-offs between model size, generation behavior, and task structure in multimodal document understanding. Larger models, such as Qwen2.5-VL 72B and Gemma3 27B, consistently outperform smaller counterparts in both language alignment and structural accuracy. While Qwen models tend to favor recall, often producing more insertions, Gemma exhibits more conservative generation with fewer deletions, particularly in high-complexity tasks such as document parsing.

These patterns suggest that model scaling alone does not uniformly reduce all error types; rather, model-family-specific tendencies (e.g., overgeneration vs. omission) influence downstream performance. Moreover, code-switching behavior appears strongly tied to parameter count, with larger models demonstrating improved language consistency. Overall, these findings emphasize the importance of balancing precision and recall depending on task demands, and highlight the need for further refinement in hallucination control for generative vision-language models.

\subsection{LLM-as-Judge vs. Traditional Metrics}

Building upon the preceding evaluation results, we next compare traditional automatic metrics with LLM-based evaluators to better understand their alignment and limitations.

Specifically, we employed GPT-4o-mini to compare model predictions against reference answers and assign scores on a 0–5 scale, averaged across samples. This evaluation was applied to the outputs of the top four models—Gemini 2.5 Pro, GPT-4o, Qwen2.5-VL-72B, and Claude Sonnet 4.

The results show that the relative rankings derived from LLM-based judgments are consistent with those from standardized automatic metrics. We observe a Pearson correlation of 0.651 and a Spearman correlation of 0.559 between LLM-as-Judge scores and traditional metrics, suggesting moderate alignment. These findings indicate that LLM-based evaluation can provide complementary insights, particularly for open-ended generative tasks where semantic adequacy is not fully captured by token-level metrics.

Nevertheless, the choice of which LLM to serve as the “judge” remains an open question requiring further systematic analysis. Considering reproducibility and cost, ThaiOCRBench currently relies on traditional automatic metrics, while LLM-based evaluation is left as a direction for future work.


\section{Conclusion}

In this work, we present ThaiOCRBench, the first comprehensive multi-task benchmark for Thai-language vision-language understanding. The benchmark consists of 2,808 human-annotated samples spanning 13 diverse tasks, designed to evaluate VLM performance on text-rich visual content in a low-resource language context.

Our zero-shot evaluation of state-of-the-art models demonstrates a clear performance disparity between proprietary and open-source systems. Gemini 2.5 Pro achieves the highest overall performance, consistently leading across most tasks. Among open-source models, Qwen2.5-VL 72B emerges as the strongest performer, though a notable performance gap remains.

Through qualitative error analysis, we identify three prominent failure modes in open-source models: language bias and code-switching, structural mismatch, and hallucinated or incorrect content. These insights underscore the need for further research and model adaptation to better support Thai-language document understanding.

ThaiOCRBench provides a standardized, task-diverse evaluation framework and establishes strong baselines for future research. We hope this benchmark will catalyze progress in developing more robust, inclusive, and linguistically aware vision-language models for Thai.

\section*{Limitations}

While ThaiOCRBench represents a significant step toward comprehensive evaluation of Thai vision-language tasks, several limitations remain. Although the dataset was explicitly designed to capture cultural specificity and long-tail document characteristics, the scale of 2,808 annotated samples may still fall short of fully representing the diversity of real-world document types. Expanding coverage across additional document genres and regional variations remains an important direction for future work.

Second, our evaluation relies primarily on conventional metrics such as BLEU, ANLS, and Tree Edit Distance. While these provide standardized measures, they may not fully capture semantic or contextual correctness. Recent approaches employing large language models as evaluators (LLM-as-Judges) offer a promising alternative for more nuanced, task-aware assessments, particularly in generative or open-ended settings.

Third, our composite metric BMFL, following the OCRBench v2 design, is computed as the unweighted average of BLEU, METEOR, F1, and normalized edit distance (NLS). This design aims to balance lexical, semantic, and character-level fidelity while maintaining reproducibility and interpretability. However, unweighted averaging may not fully capture the heterogeneous difficulty across task types. A more sophisticated multi-objective or weighted formulation would require a dedicated sensitivity study and remains an interesting avenue for future work.

Fourth, our experiments are conducted exclusively under a zero-shot setting to assess the out-of-the-box generalization capabilities of current VLMs. This does not account for performance gains achievable through fine-tuning or instruction tuning, which may significantly alter outcomes, especially for models adapted to Thai-specific data.

Lastly, ThaiOCRBench is currently monolingual, focusing solely on Thai-language content. Extending the benchmark to include multilingual and code-switched documents would enable more comprehensive evaluation of VLMs in linguistically diverse and real-world cross-lingual scenarios.

Despite these limitations, ThaiOCRBench provides a robust foundation for standardized and culturally informed evaluation of vision-language models in low-resource language settings.

\bibliography{custom}

\appendix

\section{Appendix}
\label{sec:appendix}

\subsection{Task Categories Definition}
\label{appendix:task_cat_def}

\textbf{Table parsing.} Extract tabular content from images and convert it into structured formats (e.g., Markdown or HTML), preserving both semantic relationships and layout fidelity.

\textbf{Chart parsing.} Interpret visual charts (e.g., bar, line, pie) and generate structured JSON representations using predefined entity keys that reflect underlying data semantics.

\textbf{Document parsing.} Parse full-page documents comprising heterogeneous content. Plain text is transcribed in Markdown, tables in HTML, formulas in LaTeX, and visual elements (e.g., figures and charts) are annotated using descriptive tags, enabling holistic structural understanding.

\textbf{Full-page OCR.} Transcribe all textual content present in a document image, without task-specific constraints.

\textbf{Fine-grained text recognition.} Extract targeted textual segments from specified regions within an image, emphasizing localized reading and semantic precision.

\textbf{Text recognition.} Transcribe general textual content from images, serving as a core OCR evaluation task.

\textbf{Document classification.} Assign a given document image to one of seven predefined categories based on both visual layout and textual content.

\textbf{Diagram VQA.} Answer questions grounded in diagrammatic visuals, assessing the model’s ability to jointly reason over spatial and textual components.

\textbf{Cognition VQA.} Answer image-grounded questions where the correct answer is explicitly embedded within the image, testing reading comprehension and factual grounding.

\textbf{Infographics VQA.} This task focuses on more visually complex, infographic-style inputs. Unlike Cognition VQA, it often involves dense, multi-modal layouts that require parsing of both structured and unstructured visual information.

\textbf{Key information extraction.} Extract values corresponding to provided entity keys (e.g., “Name,” “Date,” “Amount”) from densely populated documents and output them in a structured JSON format.

\textbf{Key information mapping.} Given a set of entity keys and candidate values, match each key to its corresponding value and group them into semantically meaningful clusters.

\textbf{Handwritten content extraction.} Extract and transcribe handwritten Thai text, addressing challenges associated with handwriting variability in real-world documents.

\subsection{Annotation Guideline }
\label{appendix:annotator_guideline}
This appendix outlines the step-by-step procedures and criteria followed by human annotators throughout the dataset creation process. These guidelines ensured consistency, quality, and compliance with ethical standards.

\subsubsection*{Stage 1: Data Sourcing}

\textbf{Objective:}  Collect images and ensure that they are legally and ethically usable.

\begin{itemize}
    \item \textbf{Compiling Images:} Annotators collected images either through self-captured photographs or from publicly available sources, ensuring that each image was accompanied by appropriate documentation of its license type and usage rights. 
    
    \item \textbf{Synthetic Document Generation (for PII-sensitive categories):} For sensitive categories that may involve personally identifiable information (PII) such as identification cards or certificates, images were generated using pre-defined templates and controlled data scripts. Annotators then verified that they did not contain real names, photographs, or identification numbers.
    
\end{itemize}

\subsubsection*{Stage 2: Data Annotation}

\textbf{Objective:} Accurately categorize images and remove redundant data.

\begin{itemize}
    \item \textbf{PII Sanitization:} Annotators manually reviewed each image and obscured or removed any visible faces, license plates, personal names, ID numbers or any PII information, using masking or blurring tools. To ensure accuracy and compliance, a second annotator independently reviewed the sanitization of each image.
    
    \item \textbf{Category and Metadata Tagging:} Annotators manually assigned the category to each image based on the nature of its content. In addition, annotators recorded the image source, its license type (public, licensed, synthetic, or self-taken) and optional domain tags.

    \item \textbf{Similarity Review:} Cosine similarity was computed between image embeddings to identify redundant or highly similar content. If the similarity score between two or more images exceeded a threshold of 0.95, annotators visually inspected the images. In cases where the images were found to be duplicates or near-duplicates, only the most representative sample was retained. Exceptions were made for essential images such as ID cards and official certificates, given that their high cosine similarity was necessary and due to their template-based nature.
\end{itemize}

\subsubsection*{Stage 3: Question--Answer Generation and Validation}

\textbf{Objective:} Produce question-answer pairs aligned with the image content and task objectives.

\begin{itemize}
    \item \textbf{LLM Output Triage:} Annotators manually reviewed two or three question–answer (QA) pairs generated by different large language models for each image. They then selected the pair that was most relevant, prioritizing clarity, accuracy, and alignment with the intended task objectives.

    \item \textbf{Manual Refinement:} In cases where all outputs were highly unclear, incorrect, or hallucinated, annotators manually edited the QA pairs. They were also instructed to visually inspect the image to confirm that the information referenced in the QA pair is actually present and correctly interpreted.
\end{itemize}

\subsubsection*{Stage 4: Final Quality Control}

\textbf{Objective:} Cross-validate the dataset's readiness for public release and research use.

\begin{itemize}
    \item \textbf{Coherence and Compliance Check:} A distinct set of annotators verified the logical alignment between the image, the question, and the answer. They also ensured that the image is sanitized, the question is meaningful, and the corresponding answer is directly grounded in the visual content of the image.
\end{itemize}

\begin{figure*}[t]
    \centering
    \includegraphics[width=\textwidth]{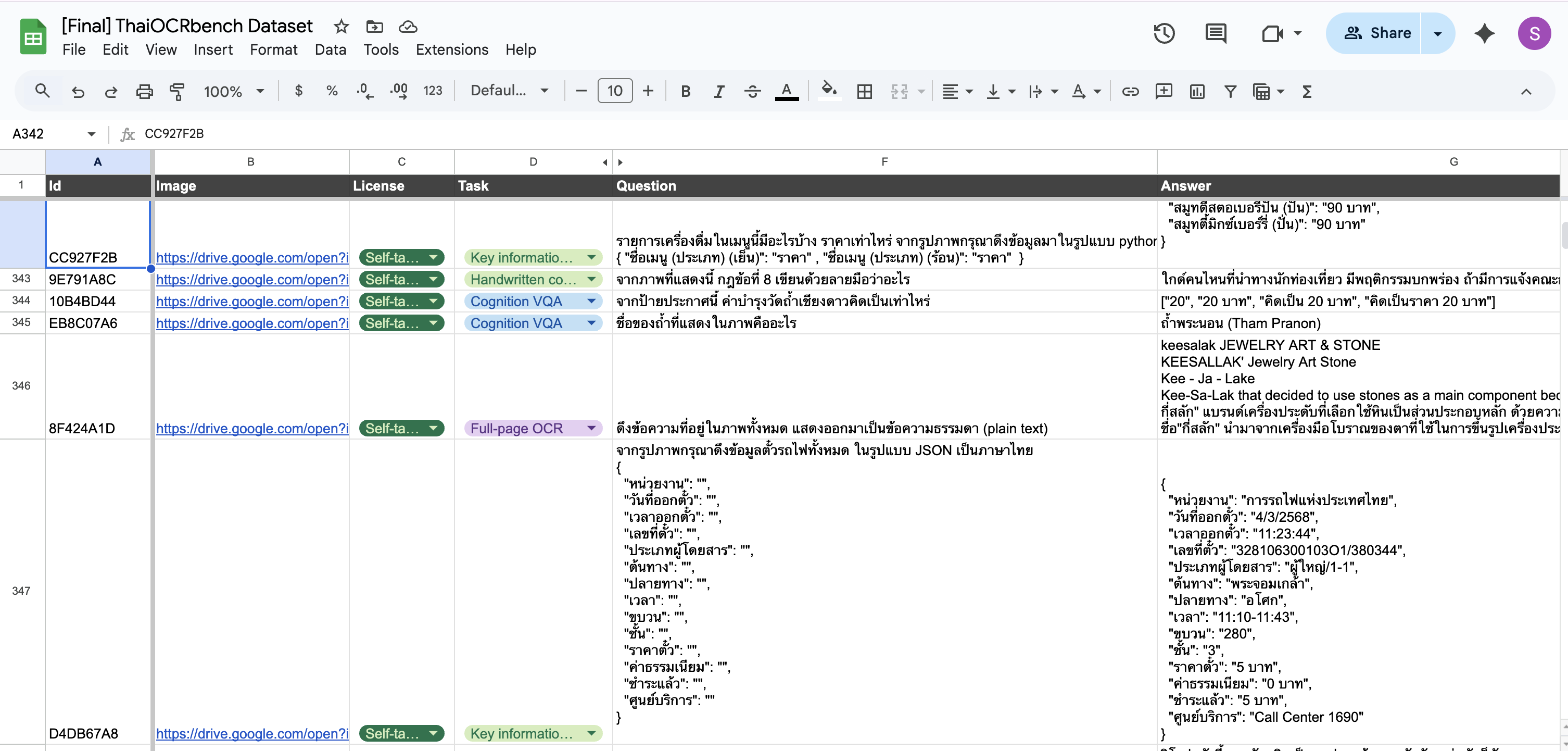}
    \caption{A screenshot of the annotation platform using Google Sheets}
    \label{fig:platform}
\end{figure*}


\begin{figure*}[t]
    \centering
    \includegraphics[width=0.65\textwidth]{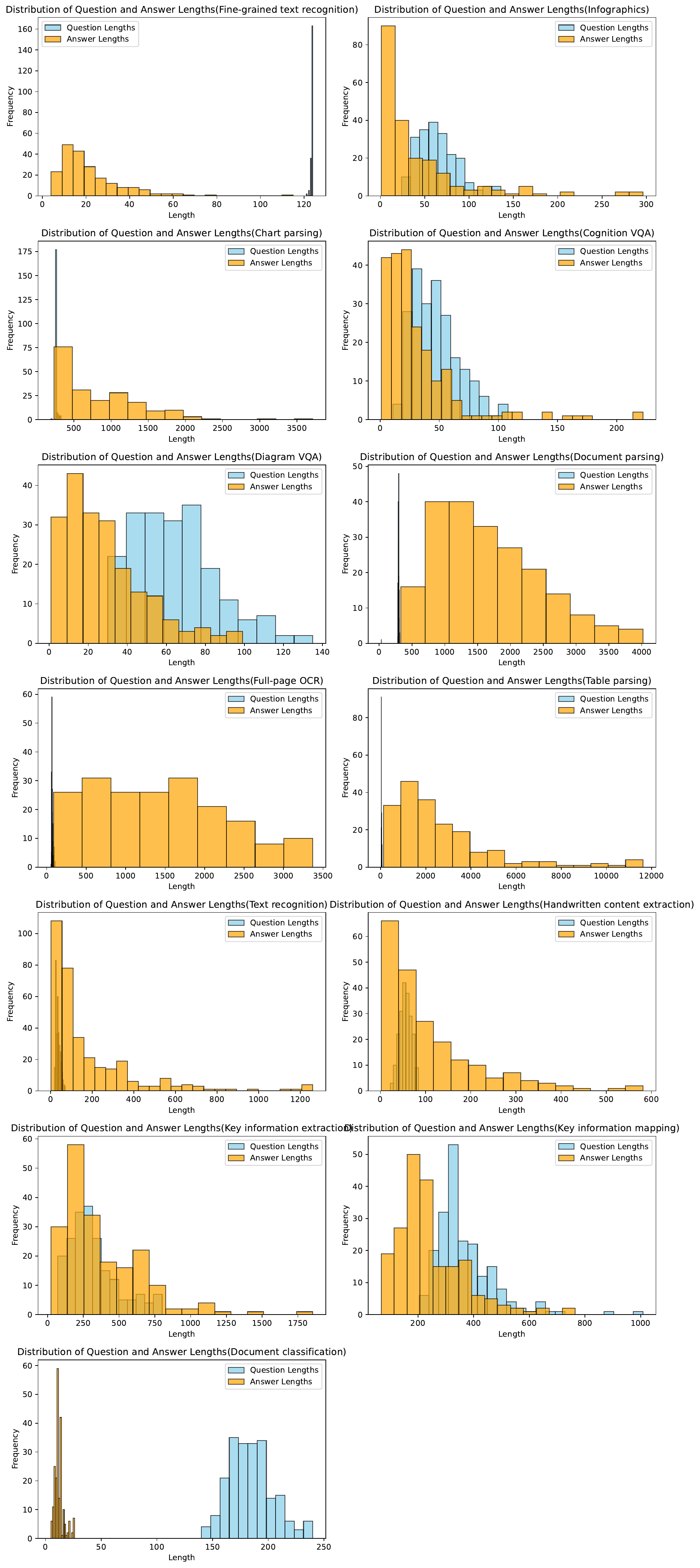}
    \caption{Comprehensive character-level statistics for both questions and answers in the ThaiOCRBench dataset.}

    \label{fig:qa_distribution}
\end{figure*}

\subsection{Dataset Samples}
\label{appendix:dataset_samples}

\begin{figure*}[t]
    \centering
    \includegraphics[width=0.9\textwidth]{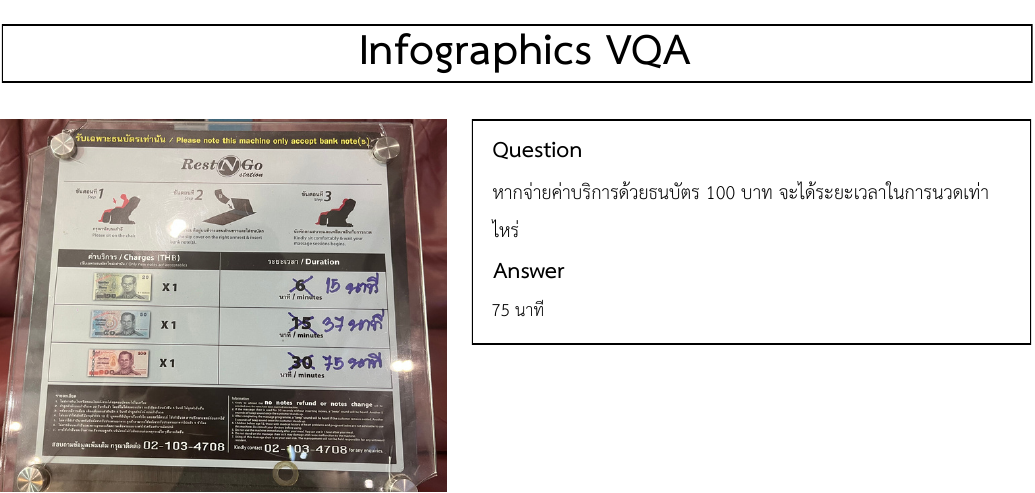}\\[1ex]
    \includegraphics[width=0.9\textwidth]{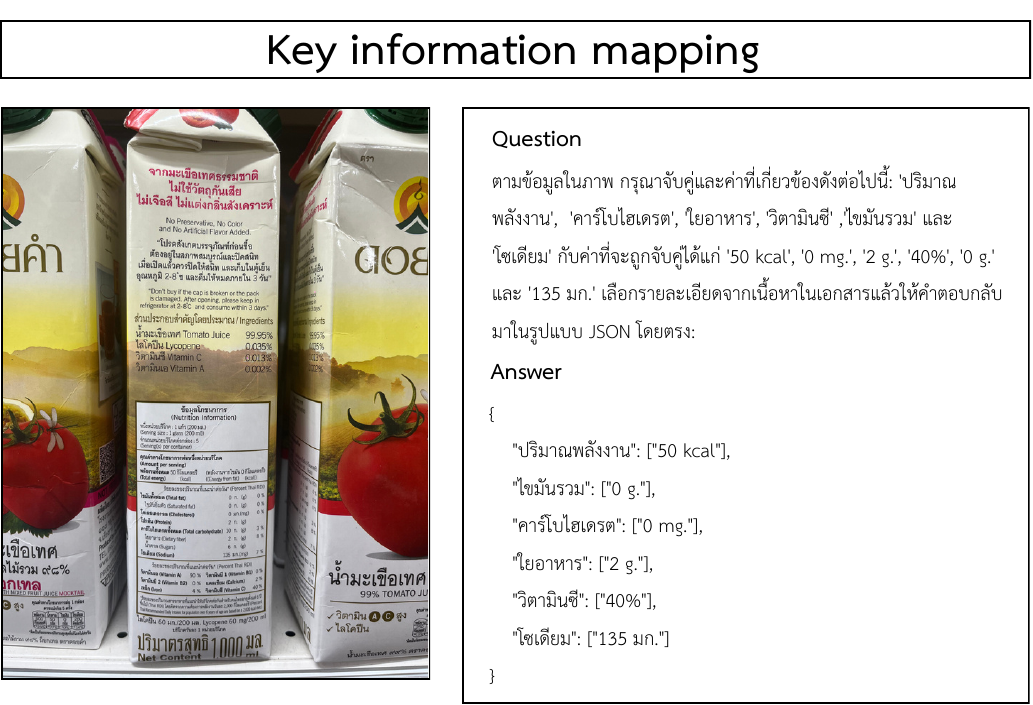}
    \caption{Examples of ThaiOCRBench: (Top) Infographics VQA, (Bottom) Key information mapping}
    \label{fig:stacked_info_and_key_map}
\end{figure*}

\begin{figure*}[t]
    \centering
    \includegraphics[width=0.9\textwidth]{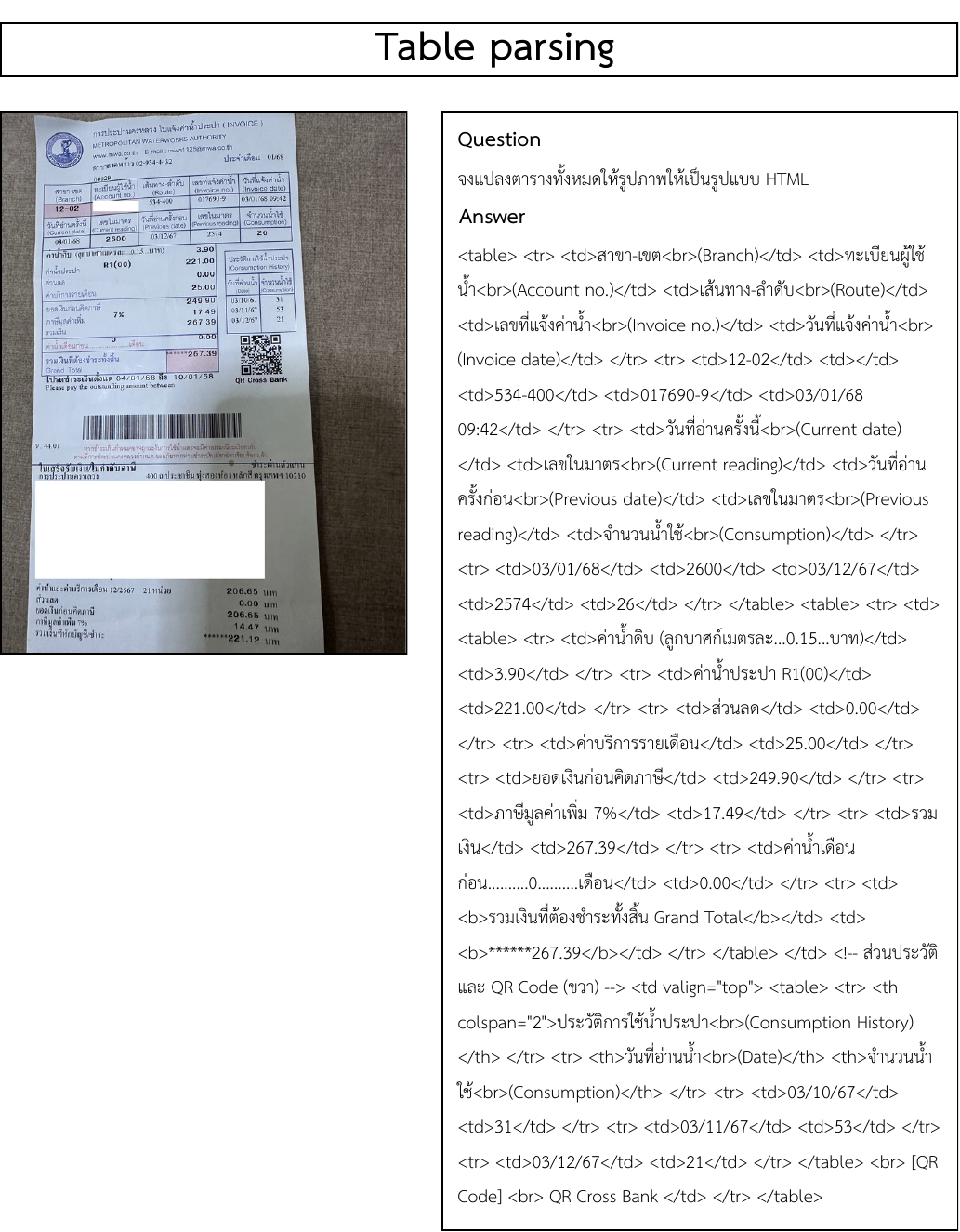}
    \caption{Examples of ThaiOCRBench: Table parsing}
    \label{fig:sample_tableparsing2}
\end{figure*}

\begin{figure*}[t]
    \centering
    \includegraphics[width=0.9\textwidth]{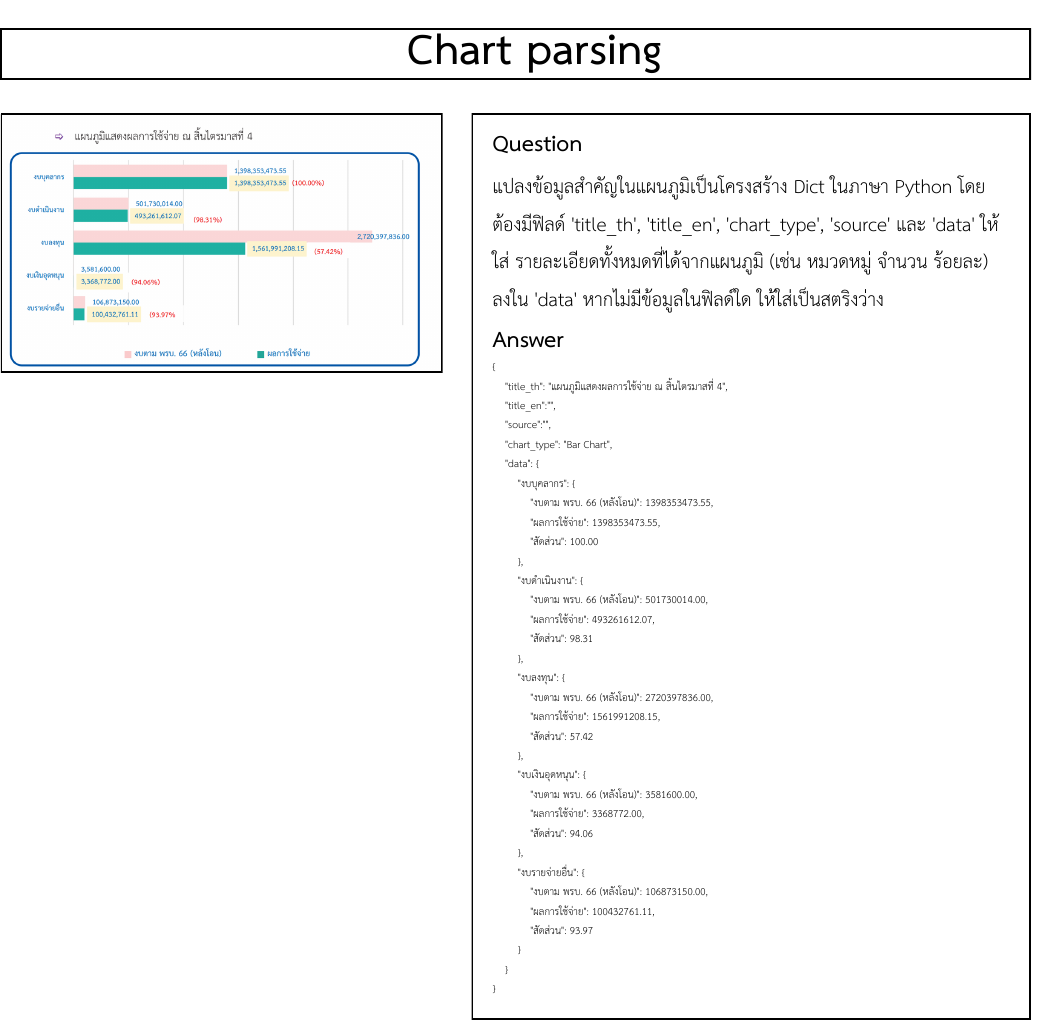}
    \caption{Examples of ThaiOCRBench: Chart parsing}
    \label{fig:sample_chartparsing2}
\end{figure*}

\begin{figure*}[t]
    \centering
    \includegraphics[width=0.9\textwidth]{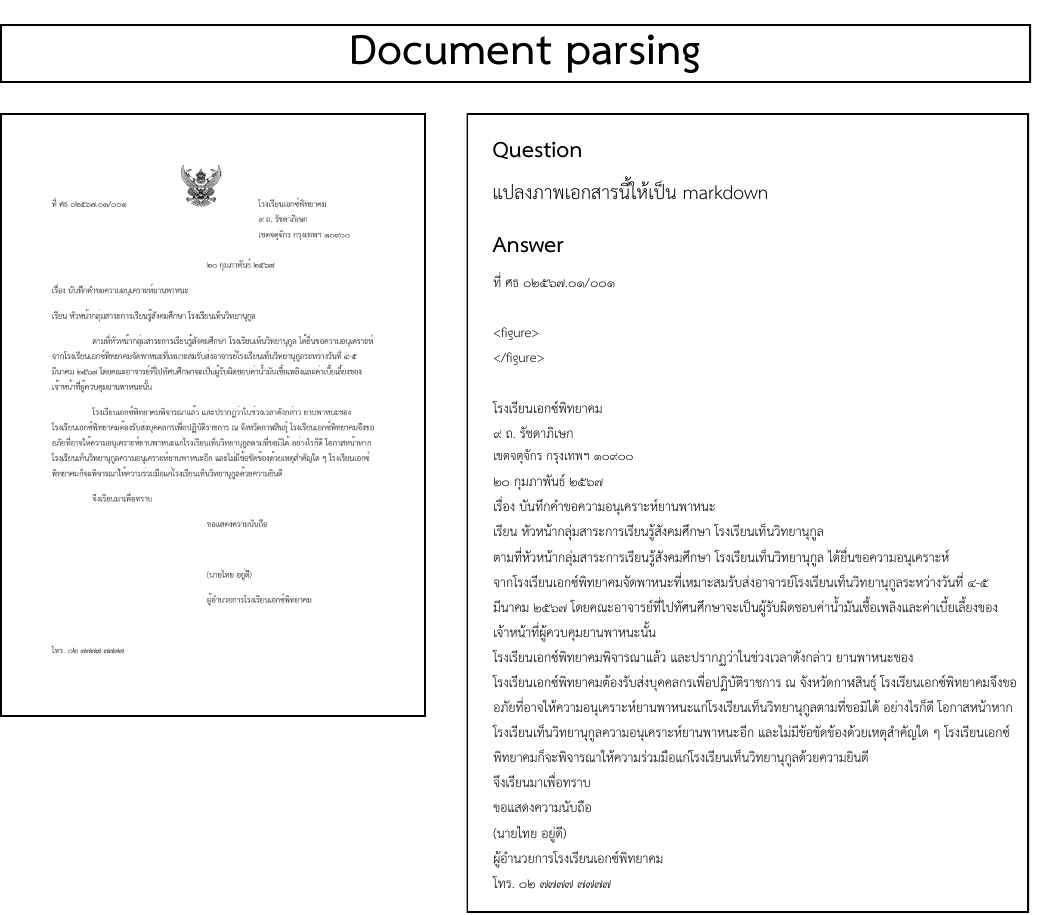}\\[1ex]
    \includegraphics[width=0.9\textwidth]{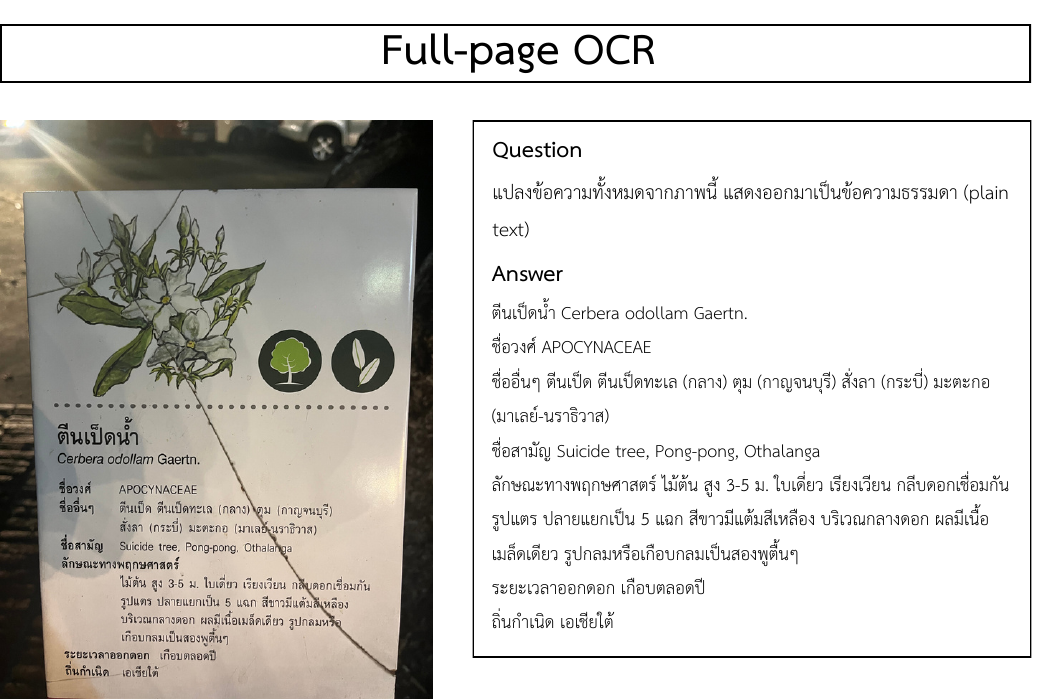}
    \caption{Examples of ThaiOCRBench: (Top) Document parsing; (Bottom) Full-page OCR}
    \label{fig:stacked_docparsing_ocr}
\end{figure*}

\begin{figure*}[t]
    \centering
    \includegraphics[width=0.9\textwidth]{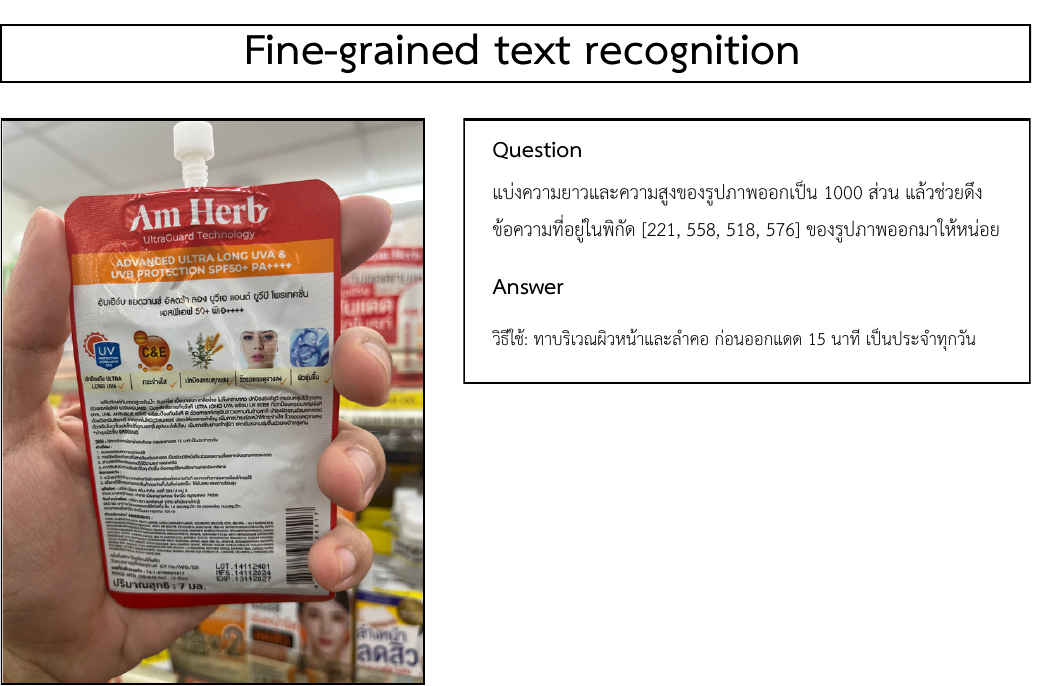}\\[1ex]
    \includegraphics[width=0.9\textwidth]{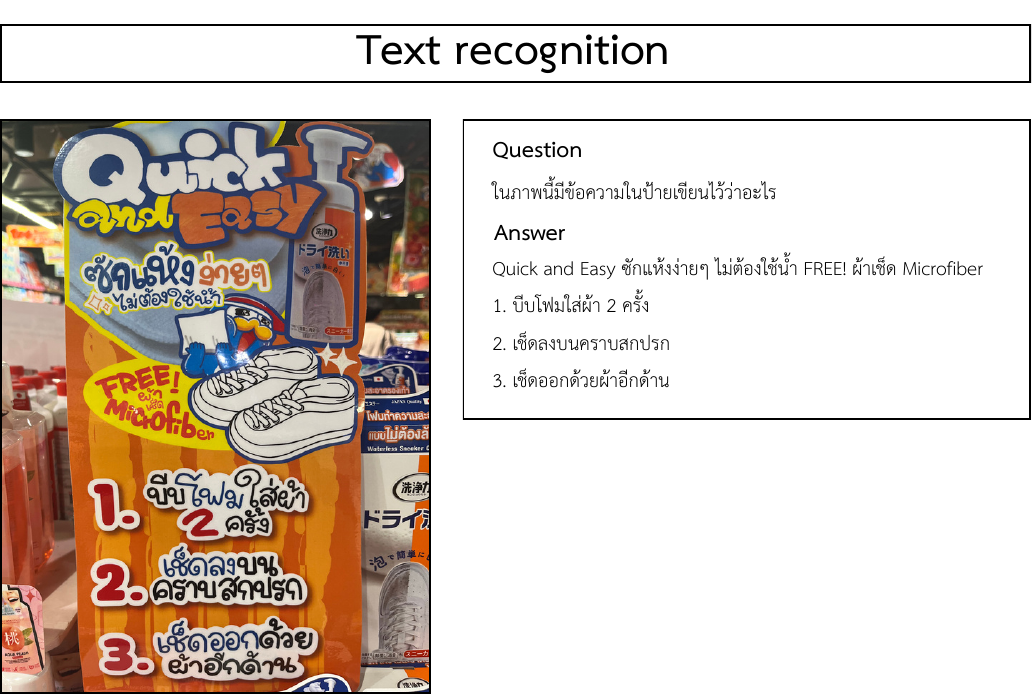}
    \caption{Examples of ThaiOCRBench: (Top) Fine-grained text recognition; (Bottom) Text recognition}
    \label{fig:stacked_textrecog}
\end{figure*}

\begin{figure*}[t]
    \centering
    \includegraphics[width=0.9\textwidth]{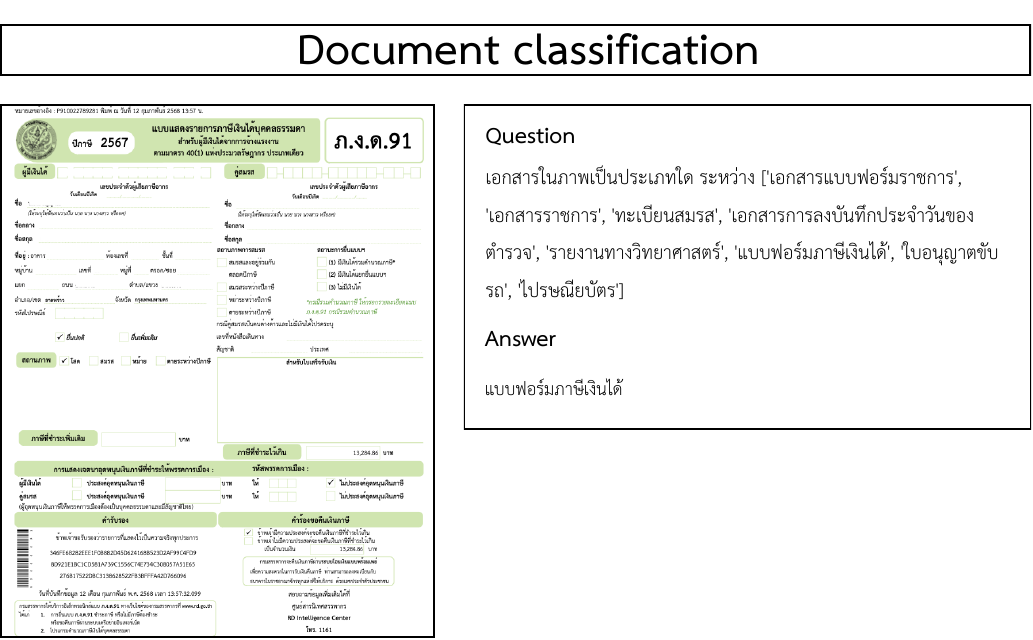}\\[1ex]
    \includegraphics[width=0.9\textwidth]{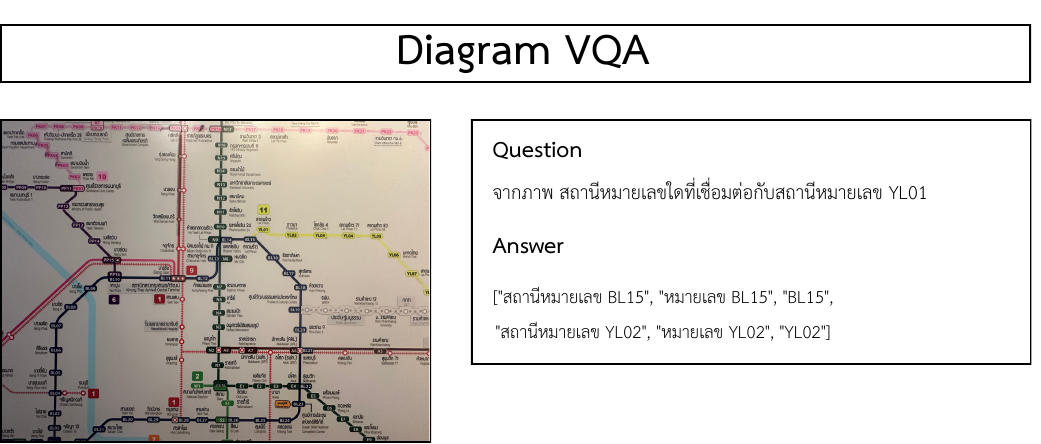}
    \caption{Examples of ThaiOCRBench: (Top) Document classification; (Bottom) Diagram VQA}
    \label{fig:stacked_docclass_vqa}
\end{figure*}

\begin{figure*}[t]
    \centering
    \includegraphics[width=0.9\textwidth]{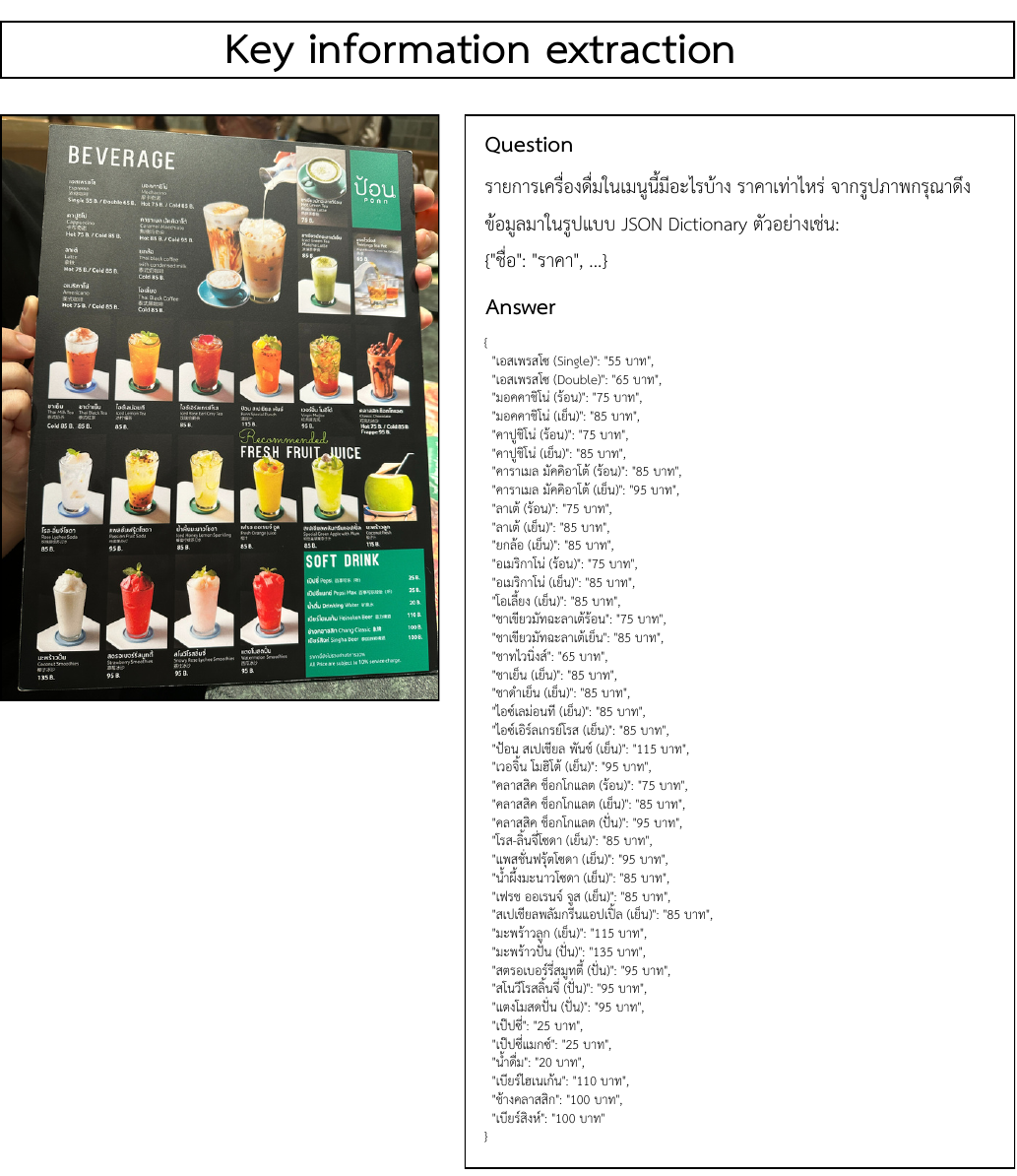}
    \caption{Examples of ThaiOCRBench: Key information extraction}
    \label{fig:sample_keyextract2}
\end{figure*}

\begin{figure*}[t]
    \centering
    \includegraphics[width=0.9\textwidth]{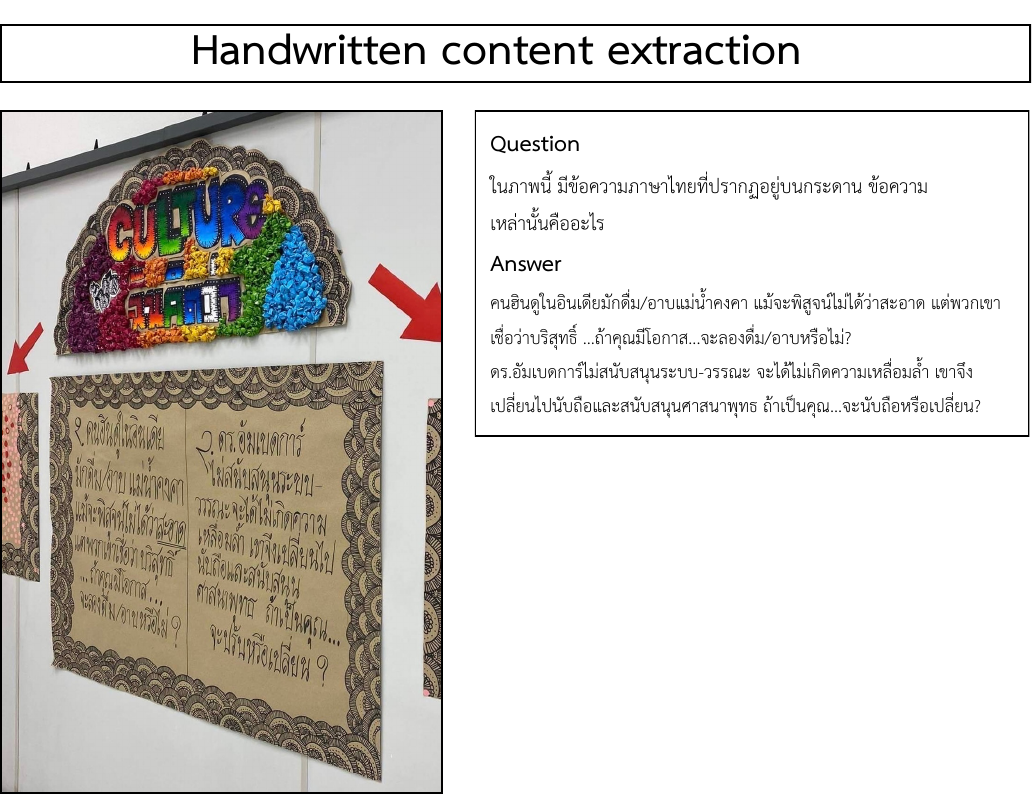}\\[1ex]
    \includegraphics[width=0.9\textwidth]{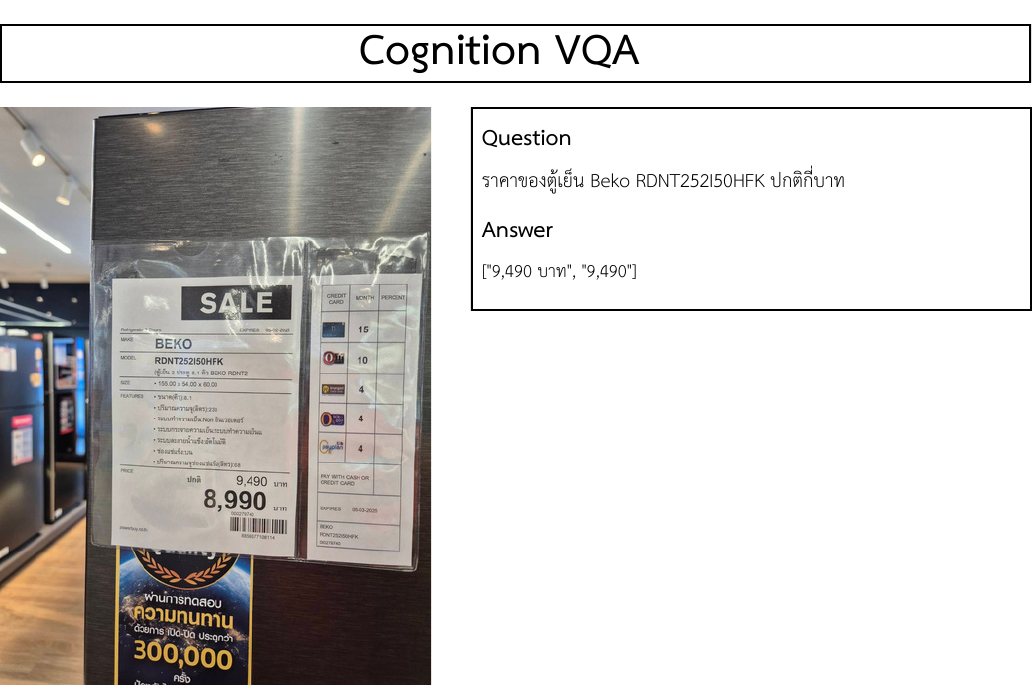}
    \caption{Examples of ThaiOCRBench: (Top) Handwritten content extraction; (Bottom) Cognition VQA}
    \label{fig:stacked_handwrite_cognition}
\end{figure*}

To illustrate the scope and diversity of our benchmark, we present representative samples for each task included in the dataset. 
Figure~\ref{fig:stacked_info_and_key_map} showcases examples from the Infographics VQA and Key information mapping tasks.  
Figure~\ref{fig:sample_tableparsing2} and Figure~\ref{fig:sample_chartparsing2} present samples of Table parsing and Chart parsing, respectively.  
Figure~\ref{fig:stacked_docparsing_ocr} includes examples of Document parsing and Full-page OCR.  
Figure~\ref{fig:stacked_textrecog} highlights Fine-grained Text recognition and Text recognition tasks.  
Figure~\ref{fig:stacked_docclass_vqa} illustrates Document classification and Diagram VQA.  
Figure~\ref{fig:sample_keyextract2} depicts a sample from the Key information extraction task.  
Lastly, Figure~\ref{fig:stacked_handwrite_cognition} presents examples of Handwritten content extraction and Cognition VQA.

These figures collectively demonstrate the wide range of document understanding tasks covered by our dataset, emphasizing its richness and complexity across both layout and semantic dimensions.

\subsection{Code-switching Prompt}
\label{appendix:code_switch_prompt}
\begin{quote}\small
\textbf{Task.} Compare the following \textbf{Answer} and \textbf{Prediction}. Determine whether there is any code switching or language bias (e.g., the Answer is in one language such as English or Thai, while the Prediction is in a different one). If the Answer and Prediction are in the \emph{same} language, that is \emph{not} code-switching.

\medskip
\textbf{Examples:}\\
\texttt{\{\{YOUR EXAMPLES FEW SHOT\}\}}

\medskip
\textbf{Answer:} \texttt{\{ANSWER\_HERE\}}\\
\textbf{Prediction:} \texttt{\{PREDICTION\_HERE\}}
\end{quote}



\end{document}